\documentclass[a4paper,fleqn,final]{cas-dc}

\usepackage[authoryear,longnamesfirst,numbers]{natbib}

\usepackage{natbib}
\setcitestyle{numbers,open={[},close={]},comma}

\usepackage[linesnumbered, ruled]{algorithm2e}
\usepackage{algpseudocode}
\usepackage{multirow}
\usepackage{arydshln}
\usepackage{setspace}
\usepackage{rotating}
\usepackage{amsthm}
\usepackage{tabularray}
\usepackage{enumitem}

\def\tsc#1{\csdef{#1}{\textsc{\lowercase{#1}}\xspace}}
\tsc{WGM}
\tsc{QE}
\newtheorem{theorem}{Theorem}
\newproof{pf}{Proof}
\newproof{pot}{Proof of Theorem \ref{thm}}

\begin{document}
\let\WriteBookmarks\relax
\def\floatpagepagefraction{1}
\def\textpagefraction{.001}

% % Short title
\shorttitle{PathMLP: Smooth Path Towards High-order Homophily}    

% % Short author
\shortauthors{J. Zhou, C. Xie, S. Gong et al.}  

\title [mode = title]{PathMLP: Smooth Path Towards High-order Homophily}

\author[1,2,3]{Jiajun Zhou}%[orcid=0000-0002-5062-4183]
\fnmark[1]
\ead{jjzhou@zjut.edu.cn}
\affiliation[1]{
    organization={Institute of Cyberspace Security, Zhejiang University of Technology}, 
    city={Hangzhou}, 
    postcode={310023},
    country={China}}
\affiliation[2]{
    organization={Binjiang Institute of Artificial Intelligence},
    city={Hangzhou},
    postcode={310056},
    country={China}}
\affiliation[3]{
    organization={College of Computer Science and Technology, Zhejiang University of Technology}, 
    city={Hangzhou}, 
    postcode={310023},
    country={China}}
\affiliation[4]{
    organization={Science and Technology on Communication Information Security Control Laboratory}, 
    city={Jiaxing}, 
    postcode={314033},
    country={China}}

\author[1,2]{Chenxuan Xie}%[orcid=0000-0002-4474-615X]
\fnmark[1]
\ead{hello.crabboss@gmail.com}

\author[1,2]{Shengbo Gong}%[orcid=0000-0001-7482-6524]
\ead{jshmhsb@gmail.com}

\author[1,2]{Jiaxu Qian}%[orcid=]
\ead{q1anjiaxu001@gmail.com}

\author[1,2]{Shanqing Yu}%[orcid=0000-0001-5170-8082]
\ead{yushanqing@zjut.edu.cn}
\cormark[1]
\cortext[1]{Corresponding author: Shanqing Yu.}

\author[1,2]{Qi Xuan}%[orcid=0000-0002-1042-470X]
\ead{xuanqi@zjut.edu.cn}

\author[1,4]{Xiaoniu Yang}%
\ead{yxn2117@126.com}

\fntext[1]{Both authors contributed equally to this research.}


\begin{abstract}
  Real-world graphs exhibit increasing heterophily, where nodes no longer tend to be connected to nodes with the same label, challenging the homophily assumption of classical graph neural networks (GNNs) and impeding their performance. Intriguingly, from the observation of heterophilous data, we notice that certain high-order information exhibits higher homophily, which motivates us to involve high-order information in node representation learning. However, common practices in GNNs to acquire high-order information mainly through increasing model depth and altering message-passing mechanisms, which, albeit effective to a certain extent, suffer from three shortcomings: 1) over-smoothing due to excessive model depth and propagation times; 2) high-order information is not fully utilized; 3) low computational efficiency. In this regard, we design a similarity-based path sampling strategy to capture smooth paths containing high-order homophily. Then we propose a lightweight model based on multi-layer perceptrons (MLP), named PathMLP, which can encode messages carried by paths via simple transformation and concatenation operations, and effectively learn node representations in heterophilous graphs through adaptive path aggregation. Extensive experiments demonstrate that our method outperforms baselines on 16 out of 20 datasets, underlining its effectiveness and superiority in alleviating the heterophily problem. In addition, our method is immune to over-smoothing and has high computational efficiency. The source code will be available in \url{https://github.com/Graph4Sec-Team/PathMLP}.
\end{abstract}

% Keywords

\maketitle

% Main text

\section{Introduction}
Graphs are employed for modeling various real-world interaction scenarios, such as social networks~\cite{social}, recommendation systems~\cite{recommendersystems} and financial transactions~\cite{fraudfinancial,zhou2022behavior}.
Recently, graph neural networks (GNNs) have gained significant attention and advancement in efficiently processing non-Euclidean graph-structured data. GNNs employ message-passing mechanisms to propagate, aggregate, and transform graph features for learning informative representations of nodes and edges, achieving outstanding performance in various tasks such as node classification and link prediction.

Classical GNNs~\cite{GCN,GAT,GraphSAGE} adhere to the homophily assumption, which states that nodes with similar features or identical labels are more likely to be connected. The phenomenon of graph homophily is prominently observed in social networks and citation networks.
For instance, individuals within social networks tend to establish connections with others who share similar interests; researchers in citation networks are inclined to reference papers within their own field~\cite{networks}. However, real-world graphs increasingly exhibit heterophily, where interconnected nodes possess dissimilar features or labels. For example, fraudulent accounts in financial networks often engage with a large number of victim (normal) accounts; similarly, scientific collaboration networks exhibit cross-disciplinary interactions~\cite{bornmann2008citation}.
These phenomena pose a challenge to the homophily assumption of classical GNNs, impeding their performance and giving rise to the \emph{heterophily problem}.

Existing studies~\cite{SNGNN,ASGNN} argue that in heterophilous scenarios, low-order information surrounding target nodes contains a higher level of noise. 
Unfiltered aggregation of low-order information can significantly interfere with the accurate characterization of node features.
To confirm this proposition, we conduct statistical experiments and obtain the following observations.

\begin{figure}[!htb]
  \centering
  \includegraphics[width=\linewidth]{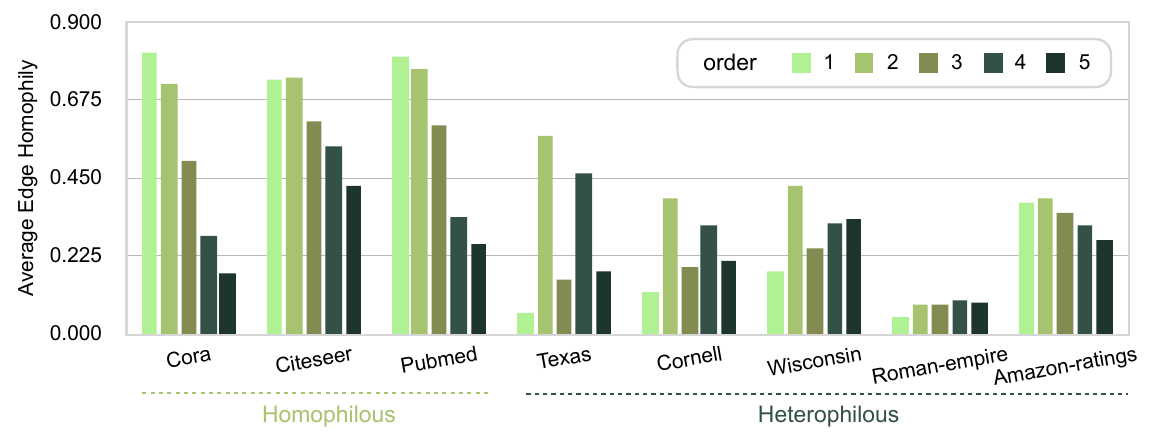}
  \caption{Average edge homophily in different order. The computation of this metric is defined in Eq.(\ref{eq: avg-edge-homo}).}
  \label{Fig: order-homo}
\end{figure}

\noindent\textbf{Observation 1:} 
We calculate the average homophily of different orders in both homophilous and heterophilous datasets, as shown in Figure~\ref{Fig: order-homo}, from which we can observe that, unlike homophilous scenarios, low-order information (especially first-order) in heterophilous graphs (e.g., \emph{Texas}, \emph{Cornell} and \emph{Wisconsin}) exhibits significantly lower levels of homophily compared to high-order information.
\emph{This inspires us to utilize higher-order information in heterophilous scenarios to alleviate the heterophily problem.}

Some classical GNNs, such as Graph Attention Networks (GAT)~\cite{GAT}, first consider the importance of different neighbors and then employ stacked convolutional layers to capture higher-order information. 
While this approach explicitly mitigates the impact of heterogeneous noise, \emph{the stacking of layers gives rise to the over-smoothing problem, wherein features from different nodes become indistinguishable with increasing number of convolutional layers,} thereby diminishing model performance.
Other approaches employ high-order message-passing mechanisms, which involve utilizing high-order information during message passing. For instance, H2GCN~\cite{H2GCN} and Mixhop~\cite{Mixhop} aggregate information from both first-order and second-order neighbors while performing message passing. GPRGNN~\cite{GPRGNN} utilizes a generalized PageRanks (GPR) weight to adaptively combine node features and high-order information after multiple propagation steps. 
However, the former models still incorporate high-order information through stacking, thereby failing to address the over-smoothing problem caused by excessive model depth. Meanwhile, the latter \emph{leverages multiple propagation steps to capture higher-order information but risks over-averaging or mixing high-order features}, thus limiting their contributions to the target node.

Path-based approaches believe that node behavior preferences are implicitly encoded in path information, and realizes message aggregation by path sampling and path feature extraction (via recurrent neural networks, i.e., RNN-like models).
For example, RAWGNN~\cite{RAWGNN} uses Node2Vec~\cite{Node2vec} to simulate breadth-first search (BFS) and depth-first search (DFS), aiming to capture both the homophilous and heterophilous information. 
PathNet~\cite{PathNet} employs entropy increase idea to guide path sampling and obtain different preference paths for different nodes. However, these approaches have notable limitations: 
1) suboptimal sampling strategies. Both strategies tend to capture structural information around target nodes rather than homophilous or heterophilous information, thus failing to effectively guide node representation learning in heterophilous scenarios; 
2) inefficiency. Each sampled path necessitates processing by RNN-like models, resulting in hugh computational costs.

To address the heterophily problem while alleviating the aforementioned limitations, we propose a PathMLP method. Specifically, we first design a similarity-based path sampling strategy to acquire smooth paths exhibiting high-order homophily for target nodes. Then, the PathMLP model can encode path information through transformation and concatenation operations, and generate highly expressive node representations in heterophilous scenarios through adaptive message aggregation. We conduct extensive experiments on 5 homophilous datasets and 15 heterophilous datasets to evaluate our methods, and the results demonstrate that our method achieves the best performance on 16 out of 20 datasets, verifying its effectiveness and superiority in alleviating the heterophily problem. Moreover, our method effectively captures higher-order information through path sampling without relying on model stacking, thus successfully circumventing the issue of over-smoothing. Furthermore, in our framework, all feature extraction, propagation, aggregation, and transformation are exclusively implemented using MLPs, thereby rendering PathMLP a lightweight and highly computationally efficient solution. 
Finally, our contributions can be summarized as follows:
\begin{itemize}[leftmargin=10pt]
  \item To the best of our knowledge, we are pioneering a novel path sampling strategy based on feature similarity to effectively address the heterophily problem by incorporating high-order information.
  \item We propose a PathMLP model that exclusively employs MLPs to efficiently encode path features through adaptive message aggregation for accurate node classification in heterogeneous scenarios.
  \item Extensive experiments on 20 benchmarks verify the effectiveness and superiority of PathMLP in alleviating heterophily problems. Meanwhile, PathMLP is not disturbed by the over-smoothing problem and has high computational efficiency.
\end{itemize}

\section{Related Work}
The graph heterophily problem has been extensively investigated in previous studies, and we classify existing research into four categories based on high-order information~\cite{H2GCN,Mixhop,GPRGNN}, graph structural information~\cite{LINKX,GloGNN,FSGNN}, paths~\cite{RAWGNN,PathNet}, and other approaches~\cite{SNGNN,FAGCN,ACMGCN,GeomGCN,gong2023NHGCN}.

High-order information is considered beneficial for alleviating heterophily in graphs. Consequently, certain methods have been designed to incorporate high-order message-passing mechanisms in order to exploit such information. H2GCN~\cite{H2GCN} and Mixhop~\cite{Mixhop} incorporate both first- and second-order neighborhood information during message-passing. GPRGNN~\cite{GPRGNN} learns GPR weights to adaptively aggregate node features and high-order information.

Several studies primarily focus on graph structure information. 
LINKX~\cite{LINKX} integrates both the adjacency matrix and node features for node classification. 
GloGNN~\cite{GloGNN} utilizes adjacency information like LINKX, but also learns correlations between nodes to aggregate information from global nodes. 
FSGNN~\cite{FSGNN} aims to select features for propagation using the adjacency matrix and investigates the impact of features of different orders on performance.

Moreover, a considerable number of research suggests that path information can effectively capture the behavioral patterns of nodes, thereby assisting in characterizing node features in heterogeneous scenarios.
GeniePath~\cite{GeniePath} proposes adaptive breadth and depth functions for selecting important nodes among first-order and higher-order neighbors. It employs an aggregation mechanism similar to GAT to obtain breadth information and subsequently utilizes an RNN-like architecture with adaptive depth to process nodes from different orders.
SPAGAN~\cite{SPAGAN} leverages the shortest path to explore latent graph structures and employs path-based attention for more effective aggregation. 
PathNet~\cite{PathNet} utilizes a maximal entropy-based random walk to capture the heterophily of neighbors and retain valuable structural information. 
RAWGNN~\cite{RAWGNN} incorporates Node2Vec~\cite{Node2vec} to simulate both BFS and DFS, capturing homophily and heterophily information respectively. 
However, the sampling strategies employed by these methods are limited to capturing only the the structural features surrounding nodes, and their computational efficiency is also constrained due to the utilization of RNN-like aggregators

Other approaches aim to address the heterophily problem in either spectral or spatial domains. 
FAGCN~\cite{FAGCN} adaptively integrates low-frequency and high-frequency signals to acquire node representations. ACMGCN~\cite{ACMGCN} adaptively aggregates the three channels of high-pass, low-pass, and identity to extract more comprehensive information and can accommodate diverse heterophilous scenarios. 
Geom-GCN~\cite{GeomGCN} updates node features by aggregating structural neighborhoods in the continuous latent space.
SNGNN~\cite{SNGNN} replaces the edge weights with node similarity and employs a topk-based strategy for selecting informative neighbors. 
GGCN~\cite{GGCN} learns new signed edge weights through structure-based and feature-based edge corrections to adjust neighbor influence. 
NHGCN~\cite{gong2023NHGCN} introduces a novel metric called neighborhood homophily to guide the separation learning of different node representations.

\section{Preliminaries}
\subsection{Notations}
An attributed graph can be denoted as $G=(V,E,\boldsymbol{X})$, where $V$ and $E$ are node set and edge set respectively, $\boldsymbol{X}\in \mathbb{R}^{|V|\times f}$ is node feature matrix. 
Here we use $|V|$, $|E|$ and $f$ to denote the number of nodes and edges, and the dimension of node feature.
The structure elements $(V, E)$ can also be denoted as an adjacency matrix $\boldsymbol{A}\in\mathbb{R}^{|V|\times|V|}$ that encodes pairwise connections between the nodes, whose entry $\boldsymbol{A}_{ij} = 1$ if there exists an edge between $v_i$ and $v_j$, and $\boldsymbol{A}_{ij} = 0$ otherwise.
Based on adjacency matrix, we can define the re-normalized affinity matrix as $\tilde{\boldsymbol{A}}_\text{sym}=\tilde{\boldsymbol{D}}^{-1/2}\tilde{\boldsymbol{A}}~\tilde{\boldsymbol{D}}^{-1/2}$, where $\tilde{\boldsymbol{A}}=\boldsymbol{A}+\boldsymbol{I}$ and $\tilde{\boldsymbol{D}}=\boldsymbol{D}+\boldsymbol{I}$.
We consider a path that starts from node $v_i$ and is within its $h$-order neighborhood as:
\begin{equation}\label{eq: pi}
    \begin{aligned}
         & p_i = [v_i,v_1,v_2,\cdots,v_j,\cdots,v_{d}]\\
         & \text{with} \quad v_j\in\mathcal{N}_i^h \cup \{v_i\},\quad 1\leq j \leq d \\
    \end{aligned}
\end{equation}
where $\mathcal{N}_i^h$ is the set of neighbors within $v_i$'s $h$-order neighborhood. 
Note that $p_i$ is a simple path if and only if no nodes and edges are visited repeatedly.

\subsection{Homophily}
Homophily refers to the tendency for nodes with similar attributes or labels to be more likely connected, which is a phenomenon commonly observed in various real-world scenarios such as social and citation networks.
In this paper, we utilize two homophily metrics as follows.

\textbf{Edge homophily}~\cite{H2GCN} refers to the proportion of edges that connect two nodes belonging to the same class (i.e., intra-class edges) in a graph:
\begin{equation}
  H_\text{edge}(G)={\left|\{(v_i,v_j)\in E\mid y_i=y_j \}\right|}/{\left|E\right|}
\end{equation}
where $y$ is the label of node. The lower $H_\text{edge}(G)$ implies more inter-class edges in the graph, i.e., stronger heterophily.
We further define the computation of the average edge homophily in the $h$-th order (mentioned in Figure~\ref{Fig: order-homo}) as follows:
\begin{equation}\label{eq: avg-edge-homo}
  \begin{aligned}
    & H_\text{edge}(G,h)={\left|\{(v_i,v_j)\in E(h)\mid y_i=y_j \}\right|}/{\left|E(h)\right|}\\
    & \text{with} \quad E(h) = \{(v_i,v_j) \mid (\boldsymbol{A}^h -\sum_{k=1}^{h-1} \boldsymbol{A}^k - \boldsymbol{I})_{ij}>0\}    
  \end{aligned}
\end{equation}

\textbf{Adjusted Homophily}~\cite{adjusted} is proposed to alleviate the issue that edge homophily metric fails to effectively measure the homophily of class-imbalanced datasets due to the influence of majority class nodes.
It further consider the relationship between edge and node degrees based on edge homophily, thereby providing a more accurate assessment of homophily in imbalanced datasets.
\begin{equation}
    H_\text{adj}(G) = \frac{H_\text{edge} - {\textstyle \sum_{c=1}^{C}}D_c^{2}/(2|E|)^2 }{1 - {\textstyle \sum_{c=1}^{C}}D_c^{2}/(2|E|)^2} 
\end{equation}
where $D_c$ denotes the sum of the degrees of all nodes belonging to class $c$, and $C$ is the number of node classes.

In this paper, we utilize edge homophily to measure all the datasets except the imbalanced two, \textit{Questions} and \textit{Tolokers}, which are measured via adjusted homophily metric.

\section{Feature Similarity-based Path Sampling}\label{sec: path}
As we know, higher-order information can be captured by increasing model depth (layer stacking), introducing high-order message passing mechanisms, path sampling, and so on. However, these existing strategies exhibit certain limitations and can be summarized as follows:
\begin{itemize}[leftmargin=10pt]
  \item \textbf{Layer Stacking:} 
  A considerable portion of GNNs exhibit a high sensitivity to the number of layers, and excessive layer stacking can lead to the over-smoothing problem.
  \item \textbf{High-order Message Passing:} 
  Such mechanisms may excessively process high-order information, thereby diminishing its expressiveness and leading to the over-smoothing problem.
  \item \textbf{Path Sampling:} 
  The prevalent path sampling strategies, such as BFS, DFS, and entropy-guided sampling, primarily concentrate on capturing structural information while often neglecting the homogeneous and heterogeneous information influenced by node features.
\end{itemize}
In this regard, we proposes a novel feature similarity-based path sampling strategy to capture high-order information, by progressively acquiring paths starting from the target node, guided by hop-by-hop similarity. 
This strategy ensures that the features along the paths do not deviate excessively from their starting nodes, yielding smooth paths containing high-order homophily information. 
The smoothness of features has been demonstrated to be advantageous for node representation learning~\cite{SNGNN}.

\subsection{Path Sampling and Observation}
Specifically, our path sampling proceeds as follows.
For a path $p_i = [v_i,v_1,v_2,\cdots,v_j,\cdots,v_{d}]$ of length $d$, the head node ($j=0$) is the target node $v_i$,
the second node ($j=1$, denoted as $v_1$) is one of the two most similar neighbors to $v_i$, 
the third node ($j=2$, denoted as $v_2$) is one of the three most similar neighbors to $v_1$, and so forth,
the fifth node ($j=4$, denoted as $v_4$) is one of the five most similar neighbors to $v_3$. 
Starting from the sixth node ($j=5$, denoted as $v_5$), each subsequent node ($j\geq 5$) is the most similar neighbor to the previous one. The sampling process stops when the path reaches the specified length $d$. 
From the above process, when $d$ is \{1, 2, 3, 4, 5, 6, $\cdots$\}, we can respectively obtain \{2 ,6, 24, 120, 120, 120, $\cdots$\} candidate paths. 
Note that the number of candidate paths no longer increases when the length of the path to be sampled exceeds 4, thereby preventing exponential growth in the number of candidate paths caused by excessively large sampling parameters.
Finally, nodes on the path can be represented as follows:
\begin{equation}\label{eq: path}
  p_i[j]\in \left\{\begin{array}{llc}
    \{v_i\} & \text { for } & j=0 \vspace{3pt}\\
    \textsf{Top}\left(j+1,  \mathcal{N}_i^1, \  \boldsymbol{x}_i^\top \cdot \boldsymbol{x}_j \right) & \text { for } & j=1 \vspace{3pt}\\
    \textsf{Top}\left(j+1,  \mathcal{N}_{j-1}^1, \  \boldsymbol{x}_{j-1}^\top \cdot \boldsymbol{x}_j \right) & \text { for } & 2 \leq j \leq 4 \vspace{3pt}\\
    \textsf{Top}\left(1,   \mathcal{N}_{j-1}^1, \ \boldsymbol{x}_{j-1}^\top \cdot \boldsymbol{x}_j \right) & \text { for } & j \geq 5 \\
    \end{array}\right.
\end{equation}
where $\boldsymbol{x}_j$ is the feature vector of node $v_j \in \mathcal{N}_i^1$ or $\mathcal{N}_{j-1}^1$, $\textsf{Top}\left(j, \ \mathcal{N}, \ \cdot\right)$ means getting the largest $j$ elements from the set $\mathcal{N}$ based on node similarity defined by inner product.

Following path sampling, each target node acquires a set of candidate paths, with the number of paths determined by the path length (i.e., \{2, 6, 24, 120\}). 
However, employing an excessive number of candidate paths for message aggregation at target nodes can lead to the issue of over-squashing~\cite{over-squashing}, particularly when dealing with long paths. 
To address this concern, we randomly sample a subset of candidate paths (controlled by parameter $N$) for subsequent message aggregation. 
Furthermore, in order to confirm the superiority of our path sampling strategy in capturing higher-order information, we also conduct statistical experiments and obtain the following observations.

\begin{figure}[!htb]
  \centering
  \includegraphics[width=\linewidth]{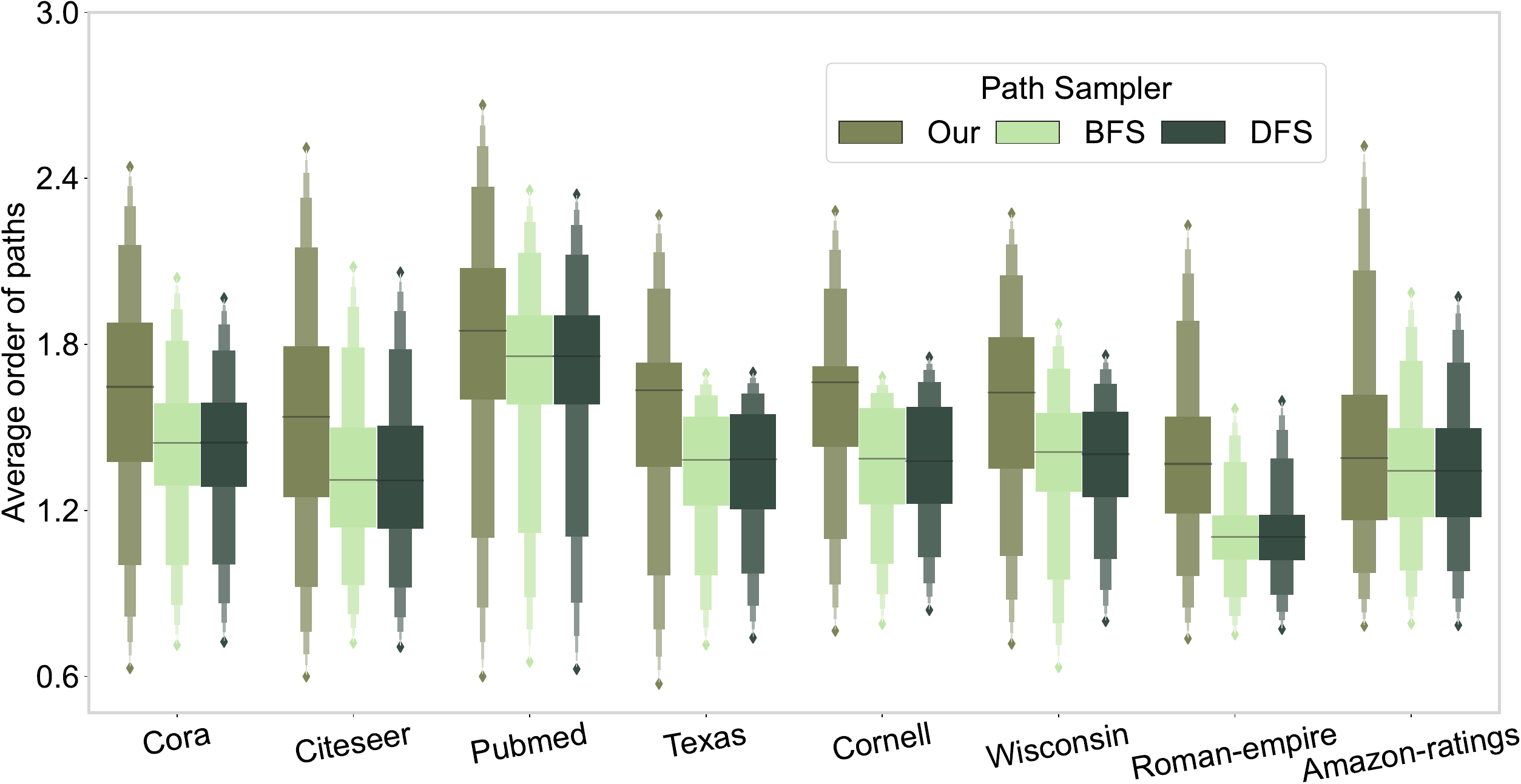}
  \caption{Statistics of the average order of candidate paths for each node obtained by different path sampling strategies.}
  \label{Fig: box-order}
\end{figure}

\noindent\textbf{Observation 2:} 
We calculate the average order of candidate paths for each node obtained by different path samplers on partial datasets, as shown in Figure~\ref{Fig: box-order}, from which we can observe that: 
1) Our feature similarity-based path sampler can effectively extract paths with higher-order information; 
2) Our sampler can extract paths with a broader range of orders.  
By combining these findings with Figure~\ref{Fig: order-homo}, we can conclude that our sampler outperforms BFS and DFS in capturing higher-order homophily.

\subsection{Theoretical Analysis of Path Sampling}
In the previous subsection, we propose to capture high-order homophily information by obtaining smooth paths through Top selection based on feature similarity sampling, which is verified in Observation 2. To further illustrate the rationality of using Top selection based on similarity sampling, we present the following theorem and prove it.
\begin{theorem}
    The discrepancy between the smooth path sampling and the expected path selection will exhibit exponential decay as the number of sampled paths increases. In other words, as the number of samples increases, the smooth paths actually sampled will approach the expected path selection results with arbitrarily high probability.
\end{theorem}
\begin{proof}
    First following the path defined by Eq.~(\ref{eq: pi}), we denote the transition probability matrix during path sampling as $\boldsymbol{S}$, where the element $s_{i,j}$ denotes the probability of sampling node $v_j$ at node $v_i$:
    \begin{equation}
      s_{i,j} = \frac{\boldsymbol{x}_i^\top \cdot \boldsymbol{x}_j}{\sum_{v \in  \mathcal{N}_i^1} \boldsymbol{x}_i^\top \cdot \boldsymbol{x}_v} 
    \end{equation}
    Thus, the sampling probability for $p_i$ can be denoted as:
    \begin{equation}
      P_{p_i}=s_{i,1}\cdot s_{1,2}\cdot... \cdot s_{j-1,j} \cdot ... \cdot s_{d-1,d}
    \end{equation}
    Considering that the set of all smooth paths starting from node $v_i$ defined by Eq.~(\ref{eq: path}) is $\mathcal{P}$, then the probability of sampling a smooth path each time is:
    \begin{equation}
        P_\text{smooth} = \sum_{p\in \mathcal{P}} P_p
    \end{equation}
    Therefore, the probability of obtaining a non-smooth path in each sampling is $1-P_\text{smooth}$. Since the events of obtaining a smooth path in each sampling are independent and identically distributed (i.i.d.), the number of smooth paths $N_\text{smooth}$ obtained in $m$ samplings follows a binomial distribution:
    \begin{equation}
        N_\text{smooth} \sim \textbf{B}(m, P_\text{smooth})
    \end{equation}
    We then define the random variable $X_i$ to represent whether the path sampled on the $i$-th occasion is a smooth path; if so, $X_i=1$, otherwise, $X_i=0$, where $i=1,2,\cdots,m$.
    The expectation $\mathbb{E}(X_i)$ of the random variable $X_i$ represents the probability of obtaining a smooth path in a single sampling, which can be given as $\mathbb{E}(X_i)=P_\text{smooth}$. 
    According to the Law of Large Numbers, as the number of samplings $m$ increases, the sample average of the random variable $\bar{X} =\frac{1}{m} {\textstyle \sum_{i=1}^{m}}  X_i$ will converge to its expected value $\mathbb{E}(\bar{X}) = \mathbb{E}(X_i) =P_\text{smooth}$ with high probability.
    To further quantify the speed and certainty of this convergence, we apply Hoeffding's Inequality~\cite{Hoeffding}:
    \begin{equation}
    \begin{array}{c}
         \text{Pr}\left(\left|\bar{X} - \mu \right|\ge \varepsilon \right)\le 2~\text{exp}(-2m\varepsilon ^2) \vspace{5pt}\\
         \Rightarrow \text{Pr}\left(\left|\frac{N_\text{smooth}}{m}  - P_\text{smooth} \right|\ge \varepsilon \right)\le 2~\text{exp}(-2m\varepsilon ^2)
    \end{array}
    \end{equation}
    where $\epsilon > 0$ is the given error margin. 
    It can be observed that, with the increase in the number of samplings, the discrepancy between smooth path sampling and expected path selection will decay exponentially, and the sampling results guided by similarity will approach the expected path selection with arbitrarily high probability.
\end{proof}

\begin{figure*}[htp]
  \centering
  \includegraphics[width=\textwidth]{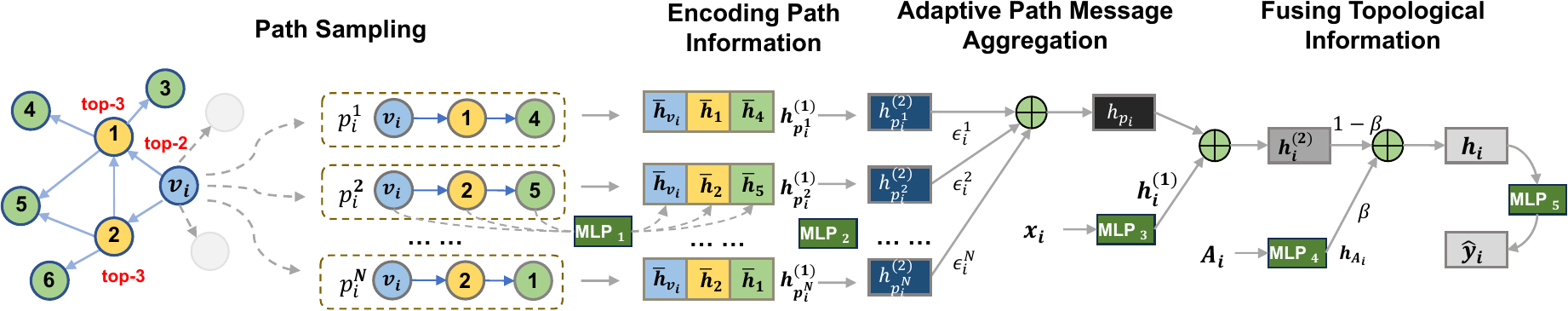}
  \caption{Illustration of the PathMLP framework.}
  \label{fig: framework}
\end{figure*}

\section{Methodology}
In this section, we will detail our PathMLP model and its variant. 
The overall framework is illustrated in Figure~\ref{fig: framework}.

\subsection{Model: PathMLP}
The aforementioned path sampling strategy can acquire smooth paths that are subsequently utilized for message aggregation, enabling the target node to fuse higher-order homophily in heterophilous scenarios. 
Previous studies~\cite{RAWGNN,PathNet} have employed RNN-like models as aggregators to process paths, which encounter two issues: 
1) the architecture of RNN-like models compresses the features along the entire path towards the target node, leading to the over-squashing problem;
2) each path requires separate processing with RNN-like models, resulting in a huge computational cost. 
To address these issues, in this paper, we propose a PathMLP model that employs only MLPs as aggregators to aggregate path information for the target nodes. 
In other words, it can encode messages carried by paths through simple transformation and concatenation operations, and adaptively aggregates path messages to target nodes.

\emph{\textbf{Step 1: Encoding Path Information.}}
After path sampling, each target node $v_i$ acquires $N$ paths $\{p_i^k \mid k=1,\cdots, N\}$ that contain high-order homophilous information. 
To encode the information of each path $p_i^k$, which belongs to $v_i$, we apply linear transformation and non-linear activation to every node on the path and then concatenate their outputs:
\begin{equation}\label{eq: 5}
  \begin{aligned}
    &\bar{\boldsymbol{h}}_v = \sigma\left(\textsf{MLP}_1\left(\boldsymbol{x}_v\right)\right)  \quad \text{for} \ \  v \in p_i^k \\
    &\boldsymbol{h}_{p_i^k}^{(1)} =\quad \parallel_{v\in p_i^k} \bar{\boldsymbol{h}}_v \\
  \end{aligned}
\end{equation}
where $\textsf{MLP}_1$ is parameterized by $\boldsymbol{W}_1\in \mathbb{R}^{f \times f'}$, $\sigma$ is the activation function, and $\parallel$ is the concatenation operation.
Here the $\textsf{MLP}_1$ can enhance the expressiveness of node features through nonlinear encoding with an activation function, and also reduces their dimensionality to prevent the exponential growth of path feature dimensions after concatenation, thereby reducing computational and memory overhead.

To further improve the expressiveness of the path features, we again apply linear transformation and nonlinear activation to $\boldsymbol{h}_{p_i^k}^{(1)}$:
\begin{equation}
  \boldsymbol{h}_{p_i^k}^{(2)} = \sigma\left(\textsf{MLP}_2\left(\boldsymbol{h}_{p_i^k}^{(1)}\right)\right)
\end{equation}
where $\textsf{MLP}_2$ is parameterized by $\boldsymbol{W}_2\in \mathbb{R}^{(1+d)\cdot f' \times f_h}$, $f_h$ is the hidden dimension.
Here $\textsf{MLP}_2$ reduces the dimensionality of the path features to align them with the encoded node features, thus optimizing computation and memory usage.

\emph{\textbf{Step 2: Adaptive Path Message Aggregation.}}
Different paths capture varying high-order features, each with a distinct impact on the characterization of the target node's representation.
To better quantify such impact while keeping the model lightweight, we introduce learnable coefficients as attention weights. This allows for the adaptive aggregation of different path features, forming the path message:
\begin{equation}
  \boldsymbol{h}_{p_i} = \sum_{k=1}^{N} \epsilon_i^k \cdot \boldsymbol{h}_{p_i^k}^{(2)}  \quad   \text{with} \ \ \sum_{k=1}^{N} \epsilon_i^k = 1
\end{equation}
where $\epsilon_i^k$ is the learnable weight for the $k$-th path of node $v_i$.
After obtaining the path message for $v_i$, we update its state:
\begin{equation}
  \begin{aligned}
    \boldsymbol{h}_i^{(1)} &= \sigma\left(\textsf{MLP}_3\left(\boldsymbol{x}_i\right)\right) \\
    \boldsymbol{h}_i^{(2)} &= \boldsymbol{h}_i^{(1)} + \boldsymbol{h}_{p_i}
  \end{aligned}
\end{equation}
where $\textsf{MLP}_3$ is parameterized by $\boldsymbol{W}_3\in \mathbb{R}^{f \times f_h}$ and is used to encode the target node.

\emph{\textbf{Step 3: Fusing Topological Information.}}
Topological information plays a crucial role in characterizing nodes, yet MLPs fail to directly and effectively capture such information. To improve the quality of node representation learning, we design a gate mechanism to control the contributions of node features and topological information:
\begin{equation}
  \begin{aligned}
    &\boldsymbol{h}_{A_i} = \sigma\left(\textsf{MLP}_4\left(\boldsymbol{A}_i\right)\right) \\
    &\boldsymbol{h}_i = \beta \cdot \boldsymbol{h}_{A_i} + (1-\beta) \cdot \boldsymbol{h}_i^{(2)}
  \end{aligned}
\end{equation}
where $\textsf{MLP}_4$ is parameterized by $\boldsymbol{W}_4\in \mathbb{R}^{|V| \times f_h}$ and is used to encode topological information, $\beta\in [0,1]$ is a trainable weight to balance the node feature and topological information of node $v_i$.

\emph{\textbf{Step 4: Model Training.}}
During model training, we employ cross-entropy as the classification loss:
\begin{equation}\label{eq: 11}
  \begin{aligned}
    \hat{\boldsymbol{y}}_{i} &= \textsf{Softmax}\left(\textsf{MLP}_5\left(\boldsymbol{h}_i\right)\right) \\
    \mathcal{L} &= -\frac{1}{|V_\text{train}|} \sum_{v_i\in V_\text{train}} \boldsymbol{y}_i \cdot \log(\hat{\boldsymbol{y}}_{i})
  \end{aligned}
\end{equation}
where $\textsf{MLP}_5$ is parameterized by $\boldsymbol{W}_5 \in \mathbb{R}^{f_h \times C}$, $C$ is the number of node classes, and $\boldsymbol{y}_i$ is the one-hot label of $v_i$. Algorithm~\ref{alg: training} shows the process of training PathMLP.

\subsection{Variant: PathMLP+}
Previous studies~\cite{SGC,deepinsight} have highlighted that the GCN-like graph convolution, as a specific case of Laplacian smoothing, can improve the smoothness between the target node and its surrounding nodes by mixing their features. 
This suggests that simply multiplying the raw features with the adjacency matrix can also provide the desired smoothness to some extent. 
Therefore, we introduce graph feature augmentation~\cite{zhou2022data} based on GCN-like graph convolution to further improve the smoothness of path features. 
Specifically, the augmentation involves initially applying GCN-like convolution to the raw features, followed by concatenating the smoothed features with the original ones, and ultimately updating the raw features:

\begin{equation}\label{eq: variant}
  \boldsymbol{X} \leftarrow \boldsymbol{X} \parallel \tilde{\boldsymbol{A}}^m_\text{sym} \boldsymbol{X}
\end{equation}
where $m\in \{1,2\}$.

\begin{algorithm}
  \SetNoFillComment
  \caption{Training PathMLP or PathMLP+.}
  \label{alg: training}
  \KwIn{Graph $G=(V,E,\boldsymbol{X})$, length of path $d$, number of paths $N$.}
  \tcc{Sampling paths}
  \For{$v_i \in K$}{
    Get all candidate paths for $v_i$ via Eq.~(\ref{eq: path}) \;
    Sampling $N$ paths for $v_i$: $\{p_i^k \mid k=1,\cdots,N\}$ \;
  }
  \tcc{Variant?}
  \If{\textit{feature augmentation}}{
    Update the raw features via Eq.~(\ref{eq: variant});
  }
  \tcc{Training model}
  Train model via Eq.~(\ref{eq: 5}) - (\ref{eq: 11}) \;
\end{algorithm}

\section{Experiments}
\subsection{Dataset}
Currently, commonly used heterophilous benchmarks mainly include \emph{Chameleon}, \emph{Squirrel}, \emph{Texas}, \emph{Cornell}, \emph{Wisconsin}, and \emph{Actor}. However, most existing studies focused on improving the effectiveness of models in alleviating the heterophily problem, while neglecting the quality of these benchmarks themselves. 
Platonov et al.~\cite{evaluate} found serious data leakage issues in \emph{Chameleon} and \emph{Squirrel}, where some nodes exhibit identical adjacency relationships and labels. Additionally, we argue that the scale of some benchmarks such as \emph{Texas}, \emph{Cornell} and \emph{Wisconsin} is rather small, which makes them susceptible to variations in the results based on different data splitting settings. In this regard, we collect benchmarks used in current research on the heterophily problem and categorize them into two groups:
\begin{itemize}[leftmargin=10pt]
  \item \textbf{Normal Group} consists of datasets that are not subject to data leakage, have a normal scale, and have undergone de-duplication. Specifically, they are \emph{Cora}, \emph{Citeseer}, \emph{Pubmed}, \emph{Cora-full}, \emph{Actor}, \emph{Chameleon-f}, \emph{Squirrel-f}, \emph{Tolokers}, \emph{Penn94}, and \emph{Electronics}. Notably, \emph{Chameleon-f} and \emph{Squirrel-f} are the de-duplicated versions of \emph{Chameleon} and \emph{Squirrel}.
  \item \textbf{Anomaly Group} consists of datasets that suffer from data leakage or have a smaller scale. Specifically, they are \emph{Chameleon}, \emph{Squirrel}, \emph{Texas}, \emph{Cornell}, \emph{Wisconsin}, \emph{NBA}, \emph{Questions}, \emph{Amazon-ratings}, and \emph{BGP}. 
  Platonov et al.~\cite{evaluate} identified data leakage issues in certain existing benchmarks and proposed new ones. However, they didn't evaluate these newly suggested benchmarks, such as \emph{Questions} and \emph{Amazon-ratings}, which also exhibit data leakage.
  Please see Appendix~\ref{app: data-leakage} for data leakage detection.
\end{itemize}
For all datasets, we consider datasets with graph homophily (the average homophily of all nodes) below 0.5 as heterophilous graphs. For detailed descriptions of these benchmarks and their data leakage status, please see Appendix~\ref{app: dataset}.

\begin{table*}[!htb]
  \setstretch{1}
  \centering
  \caption{Node classification results on real-world benchmarks (Normal Group). Boldface letters are used to mark the best results while underlined letters are used to mark the second best results. $\dagger$ indicates that the homophily is evaluated by adjusted homophily.}
  \label{tab: benchmarkA}
  \resizebox{\textwidth}{!}{
    \begin{tblr}{
      cells = {c},
      hline{1,23} = {-}{0.08em},
      hline{7,11,13,14,17,21} = {-}{dashed},
          cell{1}{13} = {r=6}{},
        }
                & Cora                         & Citeseer                     & Pubmed                       & Cora-full                    & Electronics                  & Actor                        & Chamenlon-f                  & Squirrel-f                   & Tolokers                     & Roman-empire                 & Penn94                        & \begin{sideways}Avg. ~Rank\end{sideways} \\
      \textbf{Homophily}   & 0.81                         & 0.74                         & 0.80                         & 0.57                         & 0.58                         & 0.22                         & 0.25                         & 0.22                         & 0.09                         & 0.05$\dagger$                         & 0.47                                                                 \\
      \textbf{\#Nodes}     & 2708                         & 3327                         & 19717                        & 19793                        & 42318                        & 7600                         & 890                          & 2223                         & 11758                        & 22662                        & 41554                                                                \\
      \textbf{\#Edges}     & 10556                        & 9104                         & 88648                        & 126842                       & 129430                       & 30019                        & 13584                        & 65718                        & 1038000                      & 65854                        & 2724458                                                              \\
      \textbf{\#Features}  & 1433                         & 3703                         & 500                          & 8710                         & 8669                         & 932                          & 2325                         & 2089                         & 10                           & 300                          & 4814                                                                 \\
      \textbf{\#Classes}   & 7                            & 6                            & 3                            & 70                           & 167                          & 5                            & 5                            & 5                            & 2                            & 2                            & 3                                                                    \\
      MLP       & 76.32 $\pm$ 0.99             & 72.56 $\pm$ 1.32             & 88.06 $\pm$ 0.40             & 60.12 $\pm$ 0.92             & 76.84 $\pm$ 0.39             & 37.14 $\pm$ 1.06             & 33.31 $\pm$ 2.32             & 34.47 $\pm$ 3.09             & 53.18 $\pm$ 6.35             & 65.98 $\pm$ 0.43             & 75.18 $\pm$ 0.35              & 13.4                                     \\
      GCN       & 88.41 $\pm$ 0.77             & 76.68 $\pm$ 1.00             & 88.19 $\pm$ 0.48             & 71.09 $\pm$ 0.62             & 65.52 $\pm$ 0.43             & 30.65 $\pm$ 1.06             & 41.85 $\pm$ 3.22             & 33.89 $\pm$ 2.61             & 70.34 $\pm$ 1.64             & 50.76 $\pm$ 0.46             & 80.45 $\pm$ 0.27              & 11.9                                     \\
      GraphSAGE & 87.93 $\pm$ 0.94             & 76.74 $\pm$ 1.05             & 89.03 $\pm$ 0.34             & 71.37 $\pm$ 0.52             & 76.77 $\pm$ 0.27             & 37.60 $\pm$ 0.95             & 44.94 $\pm$ 3.67             & 36.61 $\pm$ 3.06             & 82.37 $\pm$ 0.64             & 77.77 $\pm$ 0.49             & OOM                           & 8.0                                      \\
      GAT       & 87.78 $\pm$ 1.17             & 76.38 $\pm$ 1.23             & 87.80 $\pm$ 0.29             & 69.21 $\pm$ 0.52             & 65.82 $\pm$ 0.39             & 30.58 $\pm$ 1.18             & 43.31 $\pm$ 3.42             & 36.27 $\pm$ 2.12             & 79.93 $\pm$ 0.77             & 57.34 $\pm$ 1.81             & 78.10 $\pm$ 1.28              & 12.7                                     \\
      GPRGNN    & \textbf{88.83 $\pm$ 1.13}    & \underline{77.46 $\pm$ 0.77} & 89.55 $\pm$ 0.52             & \underline{71.78 $\pm$ 0.69} & 70.22 $\pm$ 0.36             & 36.89 $\pm$ 0.83             & 44.27 $\pm$ 5.23             & 40.58 $\pm$ 2.00             & 73.84 $\pm$ 1.40             & 67.72 $\pm$ 0.63             & 84.34 $\pm$ 0.29              & 6.9                                      \\
      H2GCN*    & 88.43 $\pm$ 1.50             & 77.35 $\pm$ 1.04             & 89.67 $\pm$ 0.42             & 70.98 $\pm$ 0.76             & 76.89 $\pm$ 0.47    & 37.27 $\pm$ 1.27             & 43.09 $\pm$ 3.85             & 40.07 $\pm$ 2.73             & 81.34 $\pm$ 1.16             & 79.47 $\pm$ 0.43             & 75.91 $\pm$ 0.44              & 6.5                                      \\
      RAWGNN    & 86.19 $\pm$ 1.22             & 75.79 $\pm$ 1.23             & 89.12 $\pm$ 0.47             & 65.75 $\pm$ 0.98             & 73.45 $\pm$ 0.50             & 37.30 $\pm$ 0.77             & 46.24 $\pm$ 4.07             & 37.44 $\pm$ 2.35             & 80.56 $\pm$ 0.73             & \textbf{82.19 $\pm$ 0.33}    & 74.90 $\pm$ 0.52              & 9.3        \\
      LINKX     & 83.33 $\pm$ 1.86             & 72.63 $\pm$ 0.80             & 86.99 $\pm$ 0.64             & 66.68 $\pm$ 0.96             & 75.01 $\pm$ 0.44             & 31.17 $\pm$ 0.61             & 44.94 $\pm$ 3.08             & 38.40 $\pm$ 3.54             & 77.55 $\pm$ 0.80             & 61.36 $\pm$ 0.60             & 84.97 $\pm$ 0.46              & 11.2                                     \\
      GloGNN*   & 85.16 $\pm$ 1.76             & 74.84 $\pm$ 1.35             & 89.35 $\pm$ 0.54             & 66.10 $\pm$ 1.10             & \underline{76.93 $\pm$ 0.36} & 37.30 $\pm$ 1.41             & 41.46 $\pm$ 3.89             & 37.66 $\pm$ 2.12             & 58.74 $\pm$ 13.41            & 66.46 $\pm$ 0.41             & 85.63 $\pm$ 0.27              & 9.9                                     \\
      FSGNN     & 88.39 $\pm$ 1.16             & 77.16 $\pm$ 0.89             & 89.94 $\pm$ 0.55             & 60.12 $\pm$ 0.92             & OOM                          & 37.14 $\pm$ 1.06             & 45.79 $\pm$ 3.31             & 38.25 $\pm$ 2.62             & \underline{83.87 $\pm$ 0.98} & \underline{79.76 $\pm$ 0.41}              & 83.87 $\pm$ 0.98              & 6.8                                      \\
      FAGCN     & 88.21 $\pm$ 1.20             & 76.63 $\pm$ 1.13             & 89.89 $\pm$ 0.36             & 71.61 $\pm$ 0.54             & 73.42 $\pm$ 1.61             & 37.59 $\pm$ 0.95             & 45.28 $\pm$ 4.33             & \underline{41.05 $\pm$ 2.67} & 81.38 $\pm$ 1.34             & 75.83 $\pm$ 0.35             & 79.01 $\pm$ 1.09              & 6.5                                      \\
      ACMGCN    & 88.50 $\pm$ 0.97             & 76.72 $\pm$ 0.70             & \underline{90.03 $\pm$ 0.44} & 71.59 $\pm$ 0.78             & 76.91 $\pm$ 0.27             & 36.89 $\pm$ 1.13             & 43.99 $\pm$ 2.02             & 36.58 $\pm$ 2.75             & 83.52 $\pm$ 0.87             & 79.57 $\pm$ 0.35             & 83.01 $\pm$ 0.46              & 6.1                                      \\
      GCNII*    & 87.84 $\pm$ 1.07             & 77.25 $\pm$ 0.92             & \textbf{90.21 $\pm$ 0.42}    & 70.57 $\pm$ 0.98             & 76.73 $\pm$ 0.35             & \underline{37.67 $\pm$ 1.10} & 44.66 $\pm$ 5.40             & 38.56 $\pm$ 2.88             & 83.71 $\pm$ 1.86             & 78.85 $\pm$ 0.54             & 75.20 $\pm$ 0.33              & 6.4                                      \\
      % GGCN    & 86.78 $\pm$ 1.42             & 75.89 $\pm$ 1.05             & OOM                          & OOM                          & \textbf{38.11 $\pm$ 1.04}    & 45.34 $\pm$ 3.69             & \textbf{42.43 $\pm$ 2.95}    & OOM & OOM  & OOM & OOM \\
      SNGNN*    & 88.23 $\pm$ 1.19             & 76.63 $\pm$ 0.75             & 89.02 $\pm$ 0.44             & 70.52 $\pm$ 0.74             & 72.84 $\pm$ 0.52             & 36.47 $\pm$ 0.79             & 43.76 $\pm$ 3.31             & 36.97 $\pm$ 2.70             & 76.90 $\pm$ 1.19             & 65.19 $\pm$ 0.57             & 75.22 $\pm$ 0.36              & 11.2                                     \\
      PathMLP   & 88.04 $\pm$ 1.06 & 77.13 $\pm$ 0.73 & 89.24 $\pm$ 0.48 & 70.88 $\pm$ 0.72 & \textbf{76.97 $\pm$ 0.46}  & \textbf{37.95 $\pm$ 0.73} & \underline{46.46 $\pm$ 5.20} & 40.61 $\pm$ 2.31 & 79.65 $\pm$ 0.83 & 77.74 $\pm$ 0.52 & \underline{85.97 $\pm$ 0.20}   & \underline{5.4}                          \\
      PathMLP+  & \underline{88.71 $\pm$ 0.91}   & \textbf{77.55 $\pm$ 0.92} & 89.75 $\pm$ 0.43 & \textbf{71.79 $\pm$ 0.57} & 76.08 $\pm$ 0.32 & 36.86 $\pm$ 0.79 & \textbf{46.74 $\pm$ 3.15} & \textbf{41.17 $\pm$ 3.00} & \textbf{84.25 $\pm$ 1.08} & 79.36 $\pm$ 0.46 & \textbf{86.18 $\pm$ 0.24}      & \textbf{3.5}
    \end{tblr}}
\end{table*}

\begin{table*}[!htb]
  \setstretch{0.9}
  \centering
  \caption{Node classification results on real-world benchmarks (Anomaly Group). Boldface letters are used to mark the best results while underlined letters are used to mark the second best results. $\dagger$ indicates that the homophily is evaluated by adjusted homophily.}
  \label{tab: benchmarkB}
  \resizebox{\textwidth}{!}{
    \begin{tblr}{
      cells = {c},
      cell{1}{11} = {r=6}{},
      hline{1,23} = {-}{0.08em},
      hline{7,11,13,14,17,21} = {-}{dashed},
        }
                & Chameleon                      & Squirrel                     & Questions                    & Amazon-ratings               & BGP                          & Texas                        & Cornell                        & Wisconsin                      & NBA                            & \begin{sideways}Avg. ~Rank\end{sideways} \\
      \textbf{Homophily}   & 0.23                           & 0.22                         & 0.02$^\dagger$                         & 0.38                         & 0.25                         & 0.09                         & 0.12                           & 0.19                           & 0.40                                                                 \\
      \textbf{\#Nodes}     & 2277                           & 5201                         & 48921                        & 24492                        & 63977                        & 183                          & 183                            & 251                            & 403                                                                  \\
      \textbf{\#Edges}     & 62792                          & 396846                       & 307080                       & 186100                       & 673866                       & 574                          & 557                            & 916                            & 21645                                                                \\
      \textbf{\#Features}  & 2325                           & 2089                         & 301                          & 300                          & 287                          & 1703                         & 1703                           & 1703                           & 95                                                                   \\
      \textbf{\#Classes}   & 5                              & 5                            & 2                            & 5                            & 8                            & 5                            & 5                              & 5                              & 2                                                                    \\
      MLP       & 49.76 $\pm$ 2.74               & 31.85 $\pm$ 1.44             & 54.08 $\pm$ 3.13             & 45.29 $\pm$ 0.56             & 65.20 $\pm$ 0.68             & 83.89 $\pm$ 4.50             & \underline{79.68 $\pm$ 4.80}   & 88.02 $\pm$ 4.64               & 52.22 $\pm$ 8.61              & 10.4                                      \\
      GCN       & 66.46 $\pm$ 2.89               & 46.31 $\pm$ 2.61             & 64.22 $\pm$ 1.36             & 47.52 $\pm$ 0.56             & 61.42 $\pm$ 0.81             & 61.67 $\pm$ 5.83             & 56.94 $\pm$ 5.44               & 61.76 $\pm$ 5.00               & 72.22 $\pm$ 5.46              & 11.6                                     \\
      GraphSAGE & 65.87 $\pm$ 3.01               & 47.80 $\pm$ 2.20             & 75.46 $\pm$ 1.19             & 49.20 $\pm$ 0.60             & 65.74 $\pm$ 0.81             & 83.33 $\pm$ 4.72             & 77.78 $\pm$ 4.90               & 87.25 $\pm$ 3.95               & 72.22 $\pm$ 4.44              & 6.8                                      \\
      GAT       & 66.42 $\pm$ 2.72               & 47.99 $\pm$ 2.58             & 76.43 $\pm$ 0.98             & 48.55 $\pm$ 0.76             & 64.64 $\pm$ 0.81             & 57.78 $\pm$ 8.26             & 53.33 $\pm$ 4.50               & 61.76 $\pm$ 4.16               & 65.87 $\pm$ 8.57              & 10.6                                     \\
      GPRGNN    & 66.29 $\pm$ 2.64               & 47.08 $\pm$ 1.42             & 59.55 $\pm$ 2.39             & 48.19 $\pm$ 0.92             & 64.47 $\pm$ 0.89             & 82.50 $\pm$ 5.08             & 78.89 $\pm$ 6.31               & 86.67 $\pm$ 4.22               & 70.16 $\pm$ 5.28              & 10.0                                      \\
      H2GCN*    & 54.66 $\pm$ 2.31               & 34.02 $\pm$ 1.20             & 75.85 $\pm$ 1.41             & 46.31 $\pm$ 0.44             & 65.16 $\pm$ 0.90             & 83.89 $\pm$ 6.25             & 76.94 $\pm$ 5.56               & 86.86 $\pm$ 2.78               & 72.06 $\pm$ 5.35              & 9.7                                      \\
      RAWGNN    & 53.71 $\pm$ 4.43               & 37.44 $\pm$ 1.73             & 74.05 $\pm$ 0.94             & 47.82 $\pm$ 0.65             & 64.95 $\pm$ 0.71             & 76.11 $\pm$ 3.97             & 73.89 $\pm$ 7.43               & 82.75 $\pm$ 3.50               & 72.54 $\pm$ 5.65              & 11.2 \\
      LINKX     & 73.93 $\pm$ 2.94               & 65.88 $\pm$ 1.06             & 75.07 $\pm$ 1.61             & \underline{52.68 $\pm$ 0.44} & 65.21 $\pm$ 0.69             & 68.33 $\pm$ 6.95             & 63.06 $\pm$ 8.29               & 61.18 $\pm$ 5.90               & 73.02 $\pm$ 4.10              & 7.7                                      \\
      GloGNN*   & 74.04 $\pm$ 2.92               & 45.73 $\pm$ 4.25             & OOM                          & 52.46 $\pm$ 0.81             & OOM                          & 81.94 $\pm$ 6.96             & 79.44 $\pm$ 4.57               & \underline{88.04 $\pm$ 2.84}   & 71.27 $\pm$ 8.07              & 8.0                                      \\
      FSGNN     & 65.12 $\pm$ 2.93               & 46.86 $\pm$ 1.45             & \underline{77.15 $\pm$ 1.20} & 50.82 $\pm$ 0.76             & OOM                          & 83.89 $\pm$ 5.52             & 79.17 $\pm$ 7.55               & 87.45 $\pm$ 4.73               & 70.63 $\pm$ 3.68              & 7.1                                      \\
      FAGCN     & 58.84 $\pm$ 3.54               & 39.13 $\pm$ 1.72             & 76.39 $\pm$ 1.83             & 47.69 $\pm$ 0.77             & 65.15 $\pm$ 1.19             & 83.61 $\pm$ 5.31             & \underline{79.64 $\pm$ 6.86}               & 87.45 $\pm$ 3.48               & 70.48 $\pm$ 4.38              & 8.2                                      \\
      ACMGCN    & 68.26 $\pm$ 2.47               & 54.11 $\pm$ 1.28             & 72.78 $\pm$ 1.51             & 49.06 $\pm$ 0.32             & 65.83 $\pm$ 1.43             & 84.17 $\pm$ 6.55             & 79.44 $\pm$ 6.03               & 87.45 $\pm$ 3.94               & 70.48 $\pm$ 5.41              & 5.8                                      \\
      GCNII*    & 59.16 $\pm$ 2.96               & 43.30 $\pm$ 1.25             & 74.16 $\pm$ 0.76             & 47.98 $\pm$ 0.83             & 65.62 $\pm$ 1.01             & \underline{85.00 $\pm$ 6.44}             & 78.06 $\pm$ 4.23               & 86.86 $\pm$ 2.78               & 70.48 $\pm$ 5.25              & 8.6                                      \\
      % GGCN    & 56.68 $\pm$ 3.31               & 35.16 $\pm$ 1.71             & OOM                          & OOM                          & OOM                          & 84.72 $\pm$ 5.28             & \textbf{80.83 $\pm$ 4.80}      & 87.84 $\pm$ 3.56               & 68.10 $\pm$ 7.12                               \\
      SNGNN*    & 62.75 $\pm$ 2.63               & 34.99 $\pm$ 1.42             & 75.42 $\pm$ 0.72             & 47.35 $\pm$ 0.88             & 64.44 $\pm$ 1.22             & 64.44 $\pm$ 6.78             & 54.72 $\pm$ 8.39               & 67.06 $\pm$ 6.46               & 73.17 $\pm$ 4.46              & 11.6                                     \\
      PathMLP   & \textbf{74.40 $\pm$ 2.03} & \underline{66.17 $\pm$ 1.80} & 74.50 $\pm$ 1.12 & 52.66 $\pm$ 0.60 & \underline{66.10 $\pm$ 0.67} & \textbf{85.28 $\pm$ 5.56} & 79.44 $\pm$ 5.43 & \textbf{88.43 $\pm$ 2.84} & \underline{73.49 $\pm$ 5.75}   & \textbf{2.7}                             \\
      PathMLP+  & \underline{74.04 $\pm$ 2.66}   & \textbf{66.30 $\pm$ 1.63}  & \textbf{77.21 $\pm$ 1.13} &\textbf{52.76 $\pm$ 0.60}  & \textbf{66.32 $\pm$ 0.94} & 82.50 $\pm$ 9.17 & 75.28 $\pm$ 7.91 & 84.51 $\pm$ 2.52 & \textbf{74.76 $\pm$ 4.20}      & \underline{4.2}
    \end{tblr}}
\end{table*}

\subsection{Baselines}
We compare our PathMLP and PathMLP+ with 14 various baselines, which can be categorized into five groups: 
1) Classical GNNs: MLP~\cite{MLP}, GCN~\cite{GCN}, GraphSAGE~\cite{GraphSAGE} and GAT~\cite{GAT}; 
2) High-order methods: GPRGNN~\cite{GPRGNN} and H2GCN~\cite{H2GCN}; 
3) Path-based methods: RAWGNN~\cite{RAWGNN};
4) Methods using both node feature and graph structure: LINKX~\cite{LINKX}, FSGNN~\cite{FSGNN}, GloGNN~\cite{GloGNN};
5) Other methods: FAGCN~\cite{FAGCN}, ACMGCN~\cite{ACMGCN}, GCNII~\cite{GCNII}, SNGNN~\cite{SNGNN}.
% Refer to Appendix~\ref{app: parameter} for details on baselines.

\begin{table*}[!htb]
  \centering
  \caption{Impact of path sampling strategies on PathMLP.}
  \label{tab: path}
  \resizebox{\textwidth}{!}{
  \begin{tblr}{
    cells = {c},
    cell{2}{12} = {r=7}{},
    % vline{2,12} = {-}{},
    hline{1-2,9} = {-}{},
    hline{5-6} = {1-11}{},
  }
  Path Sampler  & Cora                  & Citeseer              & Pubmed                & Cora-full             & Actor                 & Chameleon-f           & Squirrel-f            & Tolokers              & Roman-empire         & Penn94                & Count                     \\
  Ours          & \textbf{88.22 $\pm$ 0.46} & 76.32 $\pm$ 0.71          & \textbf{88.22 $\pm$ 0.46} & \textbf{70.59 $\pm$ 1.13} & \textbf{36.34 $\pm$ 1.02}          & \textbf{43.65 $\pm$ 3.84}          & \textbf{39.06 $\pm$ 2.08}          & \textbf{78.54 $\pm$ 1.63} & 73.53 $\pm$ 0.30          & \textbf{84.80 $\pm$ 0.35} & {Our: 14\\~\\DFS: 4\\~\\BFS: 2} \\
  DFS           & 86.97 $\pm$ 1.12          & \textbf{76.51 $\pm$ 0.93} & 88.07 $\pm$ 0.49          & 70.42 $\pm$ 0.84          & 35.34 $\pm$ 1.02                   & 43.59 $\pm$ 3.90          & 36.99 $\pm$ 2.31          & 78.54 $\pm$ 1.65          & 75.29 $\pm$ 0.42          & 84.79 $\pm$ 0.26          &                           \\
  BFS           & 87.30 $\pm$ 1.14          & 76.06 $\pm$ 0.66          & 88.17 $\pm$ 0.52          & 70.56 $\pm$ 0.93          & 35.27 $\pm$ 1.00                   & 43.52 $\pm$ 3.44          & 38.29 $\pm$ 2.40          & 78.47 $\pm$ 1.64          & \textbf{81.28 $\pm$ 0.37} & 84.78 $\pm$ 0.30          &                           \\
  Path Sampler  & Electronics           & Chameleon             & Squirrel              & Texas                 & Cornell               & Wisconsin             & NBA                   & Questions             & Amazon-ratings        & BGP                   &                           \\
  Ours          & \textbf{76.68 $\pm$ 0.39} & 69.47 $\pm$ 2.51          & 62.41 $\pm$ 1.35          & 82.50 $\pm$ 3.22          & \textbf{78.06 $\pm$ 4.43} & \textbf{86.67 $\pm$ 2.23} & \textbf{72.38 $\pm$ 6.13} & \textbf{73.54 $\pm$ 1.20} & \textbf{51.93 $\pm$ 0.57} & 64.85 $\pm$ 0.99          &                           \\
  DFS           & 76.36 $\pm$ 0.38          & \textbf{69.82 $\pm$ 2.31} & \textbf{63.43 $\pm$ 1.53} & ~81.94 $\pm$ 4.39         & 76.11 $\pm$ 5.27          & 86.56 $\pm$ 3.59          & 72.06 $\pm$ 4.44          & 73.15 $\pm$ 0.97          & 51.93 $\pm$ 0.62          & \textbf{65.27 $\pm$ 1.11} &                           \\
  BFS           & 76.39 $\pm$ 0.34          & 68.75 $\pm$ 2.07          & 62.79 $\pm$ 1.26          & \textbf{83.33 $\pm$ 6.55} & 74.44 $\pm$ 4.50          & 85.49 $\pm$ 4.05          & 71.11 $\pm$ 5.12          & 72.54 $\pm$ 0.82          & 51.92 $\pm$ 0.65          & 65.00 $\pm$ 1.08          &                           
  \end{tblr}}
\end{table*}
  
\begin{table*}[!htb]
  \centering
  \caption{Impact of path sampling strategies on PathMLP+.}
  \label{tab: pathplus}
  \resizebox{\textwidth}{!}{
  \begin{tblr}{
    cells = {c},
    cell{2}{12} = {r=7}{},
    % vline{2,12} = {-}{},
    hline{1-2,9} = {-}{},
    hline{5-6} = {1-11}{},
  }
  Path Sampler & Cora                  & Citeseer              & Pubmed                & Cora-full             & Actor                 & Chamenlon-f           & Squirrel-f            & Tolokers              & Roman-empire         & Penn94                & Count                     \\
  Our           & \textbf{88.30 $\pm$ 1.56} & 76.66 $\pm$ 1.38 & 88.67 $\pm$ 0.41 & \textbf{71.38 $\pm$ 0.65} & \textbf{35.11 $\pm$ 1.28} & \textbf{44.49 $\pm$ 3.56} & \textbf{40.65 $\pm$ 2.17} & 78.49 $\pm$ 1.08 & 76.97 $\pm$ 0.40 & \textbf{84.70 $\pm$ 0.34} & {Our: 13\\~\\DFS: 3\\~\\BFS: 4} \\
  DFS           & 87.99 $\pm$ 0.87 & 76.72 $\pm$ 1.61 & \textbf{88.80 $\pm$ 0.66} & 71.10 $\pm$ 0.77 & 34.82 $\pm$ 0.77 & 43.09 $\pm$ 3.11 & 39.44 $\pm$ 1.73 & \textbf{78.76 $\pm$ 1.02} & \textbf{82.99 $\pm$ 0.26} & 84.59 $\pm$ 0.40                           \\
  BFS           & 87.71 $\pm$ 1.46 & \textbf{76.75 $\pm$ 1.49} & 88.60 $\pm$ 0.52 & 71.06 $\pm$ 0.71 & 34.58 $\pm$ 0.89 & 43.88 $\pm$ 3.97 & 39.21 $\pm$ 2.35 & 78.64 $\pm$ 0.99 & 78.35 $\pm$ 0.46 & 84.64 $\pm$ 0.29                           \\
  Path Sampler & Electronics           & Chameleon             & Squirrel              & Texas                 & Cornell               & Wisconsin             & NBA                   & Questions             & Amazon-ratings        & BGP                   &                           \\
  Our           & \textbf{75.76 $\pm$ 0.41} & 69.65 $\pm$ 2.47 & 62.89 $\pm$ 1.73 & 75.00 $\pm$ 6.80 & \textbf{74.44 $\pm$ 4.86} & \textbf{81.37 $\pm$ 3.95} & \textbf{72.86 $\pm$ 7.76} & \textbf{74.55 $\pm$ 1.66} & \textbf{51.99 $\pm$ 0.49} & \textbf{65.59 $\pm$ 0.95}                           \\
  DFS           & 75.64 $\pm$ 0.47 & 69.62 $\pm$ 2.38 & 62.17 $\pm$ 1.26 & 76.11 $\pm$ 6.31 & 70.83 $\pm$ 5.44 & 80.39 $\pm$ 3.81 & 71.27 $\pm$ 5.10 & 68.73 $\pm$ 1.29 & 51.96 $\pm$ 0.60 & 65.51 $\pm$ 0.93                           \\
  BFS           & 75.70 $\pm$ 0.40 & \textbf{70.29 $\pm$ 1.84} & \textbf{63.44 $\pm$ 1.39} & \textbf{77.22 $\pm$ 5.04} & 72.50 $\pm$ 5.47 & 80.98 $\pm$ 2.27 & 71.75 $\pm$ 5.88 & 68.64 $\pm$ 1.36 & 51.91 $\pm$ 0.61 & 65.49 $\pm$ 1.12                           
  \end{tblr}}
\end{table*}

\subsection{Experimental Settings}
For most datasets, we use random splitting (48\% / 32\% / 20\% for training / validation / testing) which is different from~\cite{GeomGCN}. 
For \textit{Tolokers}, \textit{Questions}, \textit{Roman-empire} and \textit{Amazon-ratings}, we use the splitting setting (50\% / 25\% / 25\% for training / validation / testing) according to~\cite{evaluate}. We evaluate all methods with 10 runs and report the average test accuracies (or AUC for \textit{Questions} and \textit{Tolokers}). 
We set the optimizer to Adam, the maximum epochs to 500 with 100 epochs patience for early stopping, and the hidden dimension $f_h$ to 64. The search spaces of weight decay, learning rate and dropout rate are shared by all methods.
Refer to Appendix~\ref{app: parameter} for details on the hyper-parameter settings of our methods and baselines.

\subsection{Evaluation on Node Classification}
We evaluate the performance of PathMLP and its variants via node classification tasks on 20 benchmarks. For a fair comparison, we quantify the expressive power of all models based on their performance on the normal group.

Table~\ref{tab: benchmarkA} presents the node classification results of all methods on the normal group. We can observe that our methods achieve the best performance on 3 out of 5 homophilous graphs (\textit{Citeseer}, \textit{Cora-full}, and \textit{Electronics}), as well as on 5 out of 6 heterophilous graphs (except \textit{Roman-empire}). Moreover, PathMLP and PathMLP+ achieve the top two average performance rankings, demonstrating the effectiveness and superiority of our methods across datasets with varying levels of homophily. 
Notably, PathMLP+ demonstrates superior performance compared to PathMLP across most datasets, highlighting the effectiveness of our feature augmentation. Furthermore, classical GNNs, which adhere to the homophily assumption, perform poorly on heterophilous graphs, while other baselines demonstrate fluctuating performance across different datasets.

Table~\ref{tab: benchmarkB} presents the node classification results of all methods on the anomaly group. We can observe that our methods consistently achieve optimal performance on most datasets and rank among the top two in terms of average performance. 
We first focus on datasets with data leakage issues (\textit{Chameleon}, \textit{Squirrel}, \textit{Questions}, \textit{Amazon-ratings} and \textit{BGP}), where nodes with identical neighbors and labels are distributed across the training, validation and testing sets. 
For these datasets, methods that utilize graph structure information (LINKX, GloGNN, FSGNN and our models) will exploit the inherent shortcuts, i.e., use graph structure information ($\boldsymbol{A}$) to match neighbors of the target node without using node features. In particular, LINKX, as a simple model, exhibits superior performance on these datasets compared to models specifically tailored for heterogeneous graphs.
For these small-scale datasets (\textit{Texas}, \textit{Cornell}, \textit{Wisconsin}, and \textit{NBA}), our methods achieve superior performance with high stability, as evidenced by lower standard deviations. Meanwhile, MLP performs well on \textit{Texas}, \textit{Cornell} and \textit{Wisconsin}, beating almost all other baselines, implying that nodes in these three datasets tend to rely on their own features.

\subsection{More Analysis of Path Sampling}
Based on Observation 2, we further investigate the effectiveness and superiority of our path sampling strategy via ablation studies. 
Specifically, we replace our sampler in our framework with DFS and BFS respectively, and conduct comparison experiments on all datasets. Table~\ref{tab: path} and Table~\ref{tab: pathplus} illustrate the impact of different sampling strategies on PathMLP and its variant, respectively. 
It is evident that our sampling strategy significantly contributes to achieving optimal performance across most datasets, indicating that our sampler effectively captures more valuable high-order homophily information while also demonstrating superior versatility. Conversely, the other two samplers exhibit satisfactory results only on a limited number of datasets, highlighting their restricted versatility.

\begin{table*}[!htb]
	\centering
	\caption{Impact of attention mechanism on PathMLP.}
	% \arrayrulecolor{black}
	\label{tab: attention}
	\renewcommand{\arraystretch}{1.5}
	\resizebox{\textwidth}{!}{
	\begin{tabular}{ccccccccccccc} 
	\hline
	Attention      & Cora            & Citeseer        & Pubmed          & Cora-full       & Electronics     & Actor           & Chameleon-f     & Squirrel-f      & Tolokers       & Roman-empire    & Penn94         & Avg. Rank      \\ 
	\hline
	Ours                    & 88.04 $\pm$ 1.06          & 77.13 $\pm$ 0.73          & \textbf{89.24 $\pm$ 0.48} & \textbf{70.88 $\pm$ 0.72} & \textbf{76.97 $\pm$ 0.46} & 37.95 $\pm$ 0.73          & \textbf{46.46 $\pm$ 5.20} & 40.61 $\pm$ 2.31          & 79.65 $\pm$ 0.83          & \textbf{77.74 $\pm$ 0.52} & \textbf{85.97 $\pm$ 0.20} & \textbf{1.73}  \\
	GAT Att.                & 87.99 $\pm$ 0.73          & \textbf{77.14 $\pm$ 0.86} & 88.62 $\pm$ 0.43          & 70.59 $\pm$ 0.75          & 76.21 $\pm$ 0.38          & 37.91 $\pm$ 0.94          & 46.40 $\pm$ 3.85          & \textbf{40.79 $\pm$ 2.08} & 79.70 $\pm$ 1.32          & 75.88 $\pm$ 0.55          & 85.96 $\pm$ 0.30          & 2.36           \\ 
	Dot-Prod. Att.        & 87.65 $\pm$ 1.22          & 76.69 $\pm$ 0.67          & 88.49 $\pm$ 0.47          & 70.52 $\pm$ 0.82          & 76.12 $\pm$ 0.50          & 37.77 $\pm$ 1.24          & 45.90 $\pm$ 2.44          & 40.72 $\pm$ 2.41          & \textbf{79.78 $\pm$ 1.16} & 75.37 $\pm$ 0.44          & 85.95 $\pm$ 0.31          & 4.00           \\
	Scaled Dot-Prod. & 88.00 $\pm$ 1.15          & 76.80 $\pm$ 0.89          & 88.59 $\pm$ 0.53          & 70.59 $\pm$ 0.88          & 76.08 $\pm$ 0.34          & 37.89 $\pm$ 1.36          & 45.96 $\pm$ 2.92          & 40.20 $\pm$ 2.38          & 79.72 $\pm$ 1.16          & 75.76 $\pm$ 0.45          & 85.84 $\pm$ 0.28          & 3.64           \\
	Cosine Att.             & \textbf{88.10 $\pm$ 1.15} & 76.84 $\pm$ 0.83          & 88.53 $\pm$ 0.42          & 70.64 $\pm$ 0.75          & 76.24 $\pm$ 0.42          & \textbf{37.98 $\pm$ 1.32} & 45.45 $\pm$ 3.31          & 40.38 $\pm$ 2.11          & 79.49 $\pm$ 1.20          & 75.64 $\pm$ 0.66          & 85.93 $\pm$ 0.28          & 3.18           \\
	\hline\hline
	Runtime (s/epoch)  & Cora            & Citeseer        & Pubmed          & Cora-full       & Electronics     & Actor           & Chameleon-f     & Squirrel-f      & Tolokers       & Roman-empire    & Penn94         & Avg. Rank      \\ 
	\hline
	Ours                    & 0.8833          & 0.8224          & \textbf{1.6117} & \textbf{1.9214} & \textbf{4.4011} & 1.5641          & \textbf{0.7576} & 1.0527          & \textbf{5.3890} & \textbf{1.9813} & \textbf{3.965} & \textbf{1.73}  \\
	GAT Att.                & 0.9298          & 0.8317          & 2.0202          & 2.0881          & 4.7133          & 1.5703          & 0.7841          & 0.8644          & 9.1171         & 3.0327          & 4.0214         & 2.82           \\
	Dot-Prod. Att.        & \textbf{0.8454} & 0.8047          & 1.8118          & 2.6145          & 6.5027          & 1.5639          & 0.8009          & 1.0118          & 9.1167         & 3.6580           & 4.1087         & 3.36           \\
	Scaled Dot-Prod.  & 0.9466          & \textbf{0.7928} & 2.1578          & 2.1227          & 5.0271          & \textbf{1.5328} & 0.7834          & \textbf{0.8131} & 9.0725         & 3.1279          & 4.0727         & 2.45           \\
	Cosine Att.             & 1.0294          & 0.8548          & 2.2321          & 2.1730           & 5.0335          & 1.6431          & 0.7865          & 1.0527          & 10.4226        & 3.4166          & 4.4167         & 4.55           \\
	\hline
	\end{tabular}}
  \end{table*}

\begin{table*}
	\centering
	\caption{Ablation analysis of model architecture.}
	% \arrayrulecolor{black}
	\label{tab: ablation}
	\renewcommand{\arraystretch}{1.5}
	\resizebox{\textwidth}{!}{
	\begin{tabular}{ccccccccccccc} 
	\hline
						           & Cora                  & Citeseer              & Pubmed                & Cora-full             & Electronics           & Actor                 & Chameleon-f           & Squirrel-f            & Tolokers              & Roman-empire          & Penn94                & Avg. Rank      \\ 
	\hline          
	PathMLP                        & \textbf{88.04 $\pm$ 1.06} & 77.13 $\pm$ 0.73          & \textbf{89.24 $\pm$ 0.48} & 70.88 $\pm$ 0.72          & \textbf{76.97 $\pm$ 0.46} & 37.95 $\pm$ 0.73          & \textbf{46.46 $\pm$ 5.20} & \textbf{40.61 $\pm$ 2.31} & 79.65 $\pm$ 0.83          & \textbf{77.74 $\pm$ 0.52} & 85.97 $\pm$ 0.20          & \textbf{1.64}  \\
	w/o MLP1                       & 87.74 $\pm$ 1.06          & 76.66 $\pm$ 0.83          & 88.60 $\pm$ 0.47          & 69.63 $\pm$ 0.68          & OOM                   & 36.61 $\pm$ 1.14          & 45.00 $\pm$ 4.41          & 39.28 $\pm$ 2.40          & 79.03 $\pm$ 1.25          & 73.11 $\pm$ 0.64          & \textbf{86.16 $\pm$ 0.29} & 5.64           \\ 
	w/o X (MLP3)                   & 87.19 $\pm$ 0.89          & 75.53 $\pm$ 0.83          & 88.11 $\pm$ 0.52          & 69.22 $\pm$ 0.66          & 73.08 $\pm$ 0.35          & 37.66 $\pm$ 1.03          & 45.84 $\pm$ 3.08          & 40.29 $\pm$ 2.87          & 79.67 $\pm$ 1.12          & 73.23 $\pm$ 0.50          & 85.72 $\pm$ 0.30          & 5.36           \\ 
	w/o A (MLP4)                   & 87.30 $\pm$ 1.07          & 76.83 $\pm$ 1.16          & 88.37 $\pm$ 0.52          & 68.80 $\pm$ 0.80          & 76.04 $\pm$ 0.42          & \textbf{37.97 $\pm$ 1.34} & 45.45 $\pm$ 3.72          & 38.81 $\pm$ 2.42          & 75.96 $\pm$ 0.89          & 73.64 $\pm$ 0.58          & 74.37 $\pm$ 0.27          & 5.64           \\ 
	\hdashline   
	concat $\rightarrow$ mean      & 88.01 $\pm$ 1.33          & \textbf{77.44 $\pm$ 0.75} & 87.98 $\pm$ 0.38          & \textbf{71.15 $\pm$ 0.81} & 76.47 $\pm$ 0.38          & 37.14 $\pm$ 1.29          & 45.62 $\pm$ 2.90          & 40.00 $\pm$ 2.08          & 78.56 $\pm$ 1.02          & 69.46 $\pm$ 0.53          & 85.92 $\pm$ 0.38          & 4.18           \\
	concat $\rightarrow$ sum       & 87.78 $\pm$ 1.13          & 76.93 $\pm$ 0.79          & 88.11 $\pm$ 0.45          & 70.50 $\pm$ 0.90          & 75.06 $\pm$ 0.44          & 37.67 $\pm$ 1.20          & 45.11 $\pm$ 3.59          & 38.61 $\pm$ 2.07          & 78.54 $\pm$ 1.14          & 68.92 $\pm$ 0.82          & 85.62 $\pm$ 0.36          & 5.73           \\ 
	\hdashline      
	attention $\rightarrow$ mean   & 87.82 $\pm$ 0.86          & 76.84 $\pm$ 1.03          & 88.63 $\pm$ 0.63          & 70.59 $\pm$ 0.69          & 75.65 $\pm$ 0.53          & 37.91 $\pm$ 1.26          & 45.56 $\pm$ 3.15          & 40.27 $\pm$ 2.21          & 79.66 $\pm$ 1.14          & 75.82 $\pm$ 0.48          & 85.91 $\pm$ 0.35          & 3.36           \\
	attention $\rightarrow$ sum    & 87.47 $\pm$ 1.22          & 76.32 $\pm$ 0.86          & 88.47 $\pm$ 0.60          & 70.00 $\pm$ 1.05          & 75.03 $\pm$ 0.51          & 37.51 $\pm$ 1.54          & 46.07 $\pm$ 2.74          & 39.57 $\pm$ 3.52          & \textbf{80.05 $\pm$ 1.29} & 75.24 $\pm$ 0.54          & 85.93 $\pm$ 0.37          & 4.36           \\ 
	\hline\hline      
	PathMLP: time (s)              & \textbf{0.8833}       & \textbf{0.8224}       & \textbf{1.6117}       & \textbf{1.9214}       & \textbf{4.4011}       & \textbf{1.5641}       & \textbf{0.7576}       & \textbf{1.0527}       & 5.3890                & \textbf{1.9813}       & \textbf{3.9650}       & \textbf{1.09}  \\
	w/o MLP1: time (s)             & 1.0435                & 1.7599                & 3.0090                & 24.6700               & OOM                   & 1.6611                & 0.8960                & 1.1439                & 4.7557                & 3.1146                & 29.9800               & 1.91           \\ 
	\hdashline          
	PathMLP: memory (M)            & \textbf{2095}         & \textbf{2153}         & \textbf{2475}         & \textbf{3151}         & \textbf{4391}         & \textbf{1863}         & \textbf{2055}         & \textbf{2099}         & 2333                  & \textbf{2563}         & \textbf{3775}         & \textbf{1.09}  \\
	w/o MLP1: memory (M)           & 2689                  & 4051                  & 3885                  & 29353                 & OOM                   & 3269                  & 2391                  & 2805                  & 2263                  & 3435M                 & 33989                 & 1.91           \\
	\hline
	\end{tabular}}
\end{table*}

\subsection{Analysis of Aggregation Mechanism}
We further investigate the advantages of using learnable weights for path feature aggregation.
Specifically, we compare learnable weights with four prevalent attention mechanisms categorized as either learnable or non-learnable. The former includes GAT Attention, while the latter includes Dot-Product Attention, Scaled Dot-Product Attention, and Cosine Similarity.
We substitute the learnable weights in PathMLP with these four attention mechanisms and conduct experiments on 11 datasets from the normal group using the same experimental settings as the node classification experiments. 
Table~\ref{tab: attention} reports the average test accuracy, standard deviation and average running time.
We can observe that learnable attention mechanisms generally outperform non-learnable attention mechanisms. 
And GAT attention is the closest to our learnable weights in overall performance but requires significantly more computational resources.
The comparison of average running time reveals that the other attention mechanisms consume between 68\% and 93\% more time than our learnable weights, with the difference being more pronounced on larger datasets.
We conclude that other attention mechanisms do not significantly improve performance in this context and instead increase running time. 
Therefore, for the sake of model lightweight, we ultimately choose to use learnable weights for path feature aggregation.

\subsection{Ablation Analysis of Model Architecture}
We further conduct comprehensive ablation experiments to verify the importance and effectiveness of several key components, including the path feature encoding module (MLP1), node feature encoding module (MLP3), topological information encoding module (MLP4), path feature generation method (concatenation operation), and path message aggregation method (learnable weights), as demonstrated in Table~\ref{tab: ablation}.
Specifically, we first analyze the importance of the first three modules. The performance degradation observed upon the removal of any module underscores their importance: 1) $\textsf{MLP}_1$ enhances the expressiveness of node features; 2) The target node's own features are crucial for its representation learning; 3) The topological information surrounding the target node is crucial for its representation learning.
Moreover, we also verify the impact of $\textsf{MLP}_1$ on the model's computational efficiency. Removing $\textsf{MLP}_1$ leads to significant increases in both runtime and memory consumption, indicating its role in reducing computational overhead through dimensionality reduction.
Additionally, we experiment with replacing the path feature generation method and the path message aggregation method with mean pooling and sum pooling, respectively. The performance degradation observed upon replacing any one of these modules underscores their importance: 1) The concatenation operation can better preserve the integrity of path features; 2) The adaptive aggregation approach can utilize different path features in a more effective way.

\begin{figure*}
  \centering
  \includegraphics[width=\textwidth]{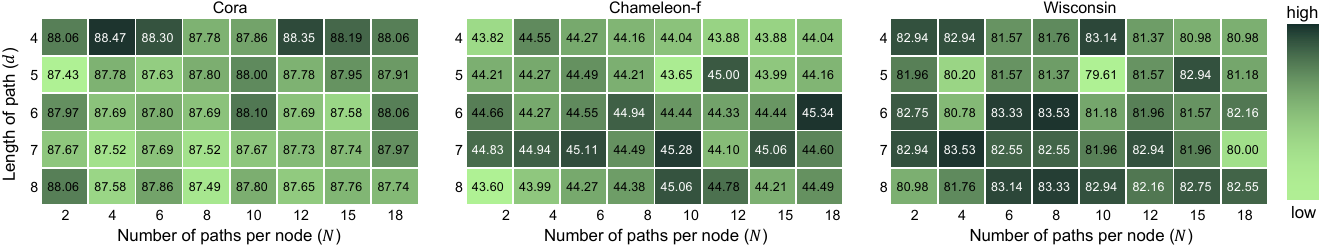}
  \caption{Impact of hyper-parameters in PathMLP+.}
  \label{Fig: hyper-parameter}
\end{figure*}

\begin{figure}[!htp]
  \centering
  \includegraphics[width=\linewidth]{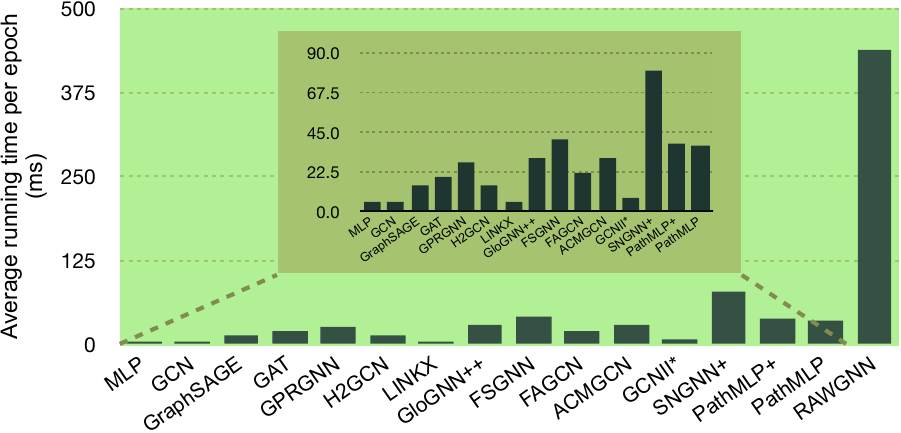}
  \caption{Average running time per epoch (ms).}
  \label{Fig: time}
\end{figure}

\begin{figure}[!htp]
  \centering
  \includegraphics[width=\linewidth]{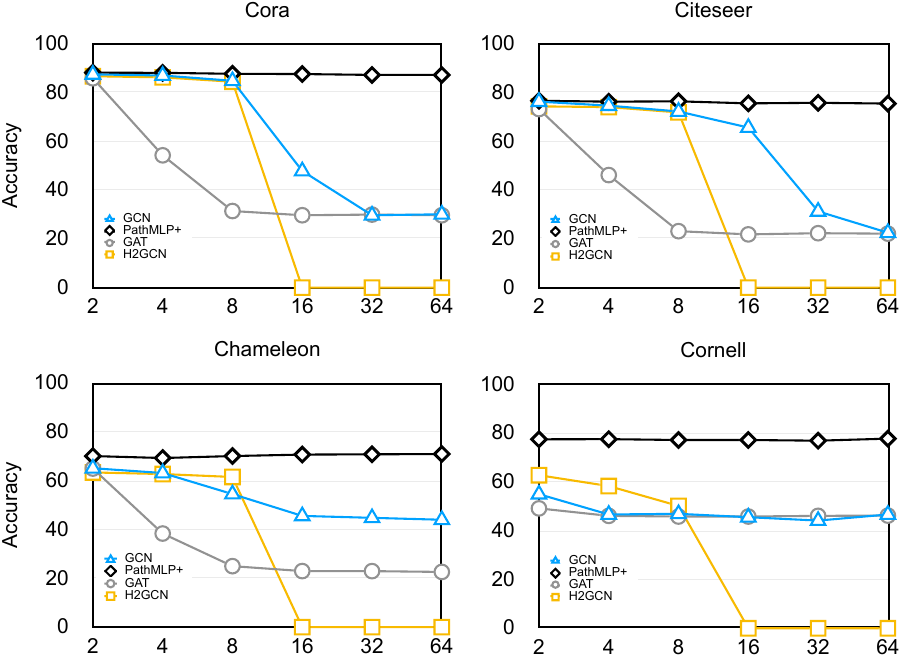}
  \caption{Performance of GNNs with increasing model depths (or path length).}
  \label{Fig: smooth}
\end{figure}

\subsection{Analysis of Hyper-parameter}
We further investigate the impact of two important hyper-parameters, i.e., the path length $d$ and the number of paths $N$, on the performance of our methods. We vary $d$ in $\{3,4,5,6,7\}$ and $N$ in $\{2,4,6,8,10,12,15,18\}$. Figure~\ref{Fig: hyper-parameter} shows the performance variation of PathMLP with different combinations of hyper-parameters, from which we can observed that: 
1) on the homophilous dataset \textit{Cora}, better performance is achieved by utilizing shorter paths; 
2) on the heterophilous datasets \textit{Chameleon-f} and \textit{Wisconsin}, better performance is attained by employing longer paths.
These phenomena are consistent with our intuition that on homophilous graphs, shorter paths suffice to capture rich low-order homophily information, whereas in heterophilous graphs, longer paths are necessary to capture more high-order information. Moreover, the impact of path quantity on the model does not show a clear pattern.

\subsection{Efficiency Analysis}
To evaluate the computational efficiency of our methods and baselines, particularly path-based methods, we conduct efficiency analysis on \textit{Pubmed}. For path-based methods (PathMLP, PathMLP+, RAWGNN), we set $N$ to 12 and $d$ to 4. 
Figure~\ref{Fig: time} shows the average running time of all methods for each epoch, from which we can observe that our methods have about 11.8 times efficiency improvement relative to RAWGNN, and this improvement grows as the number of paths ($N$) increases. This phenomenon suggests that our MLP-based methods for encoding path features are significantly more computationally efficient compared to methods utilizing RNN-like models (RAWGNN, PathNet). Furthermore, our methods demonstrate comparable computational efficiency to GloGNN++, FSGNN, ACMGCN, and GPRGNN while beating SNGNN+. In summary, PathMLP exhibits a relatively high level of computational efficiency.

\subsection{Over-smoothing Analysis}
To validate the immunity of our method against the over-smoothing problem, we conduct model depth experiments on four datasets. Specifically, we compare with classical methods (GCN, GAT) and high-order methods (H2GNN), which acquire high-order information through stacking, while our method achieves it by increasing path length. Figure~\ref{Fig: smooth} shows the performance of all methods at varying model depths (path lengths). 
It is evident that classical GNNs exhibit a rapid decline in performance as the model depth increases, indicating a significant over-smoothing phenomenon.
For H2GCN, with an increase in model depth from 2 to 8 layers, there is a gradual decline in performance, suggesting potential susceptibility to over-smoothing. When the model depth exceeds 16 layers, it leads to memory overflow within the model.
In contrast, our method shows stable performance with increasing path length, indicating its immunity to the over-smoothing problem.

\section{Conclusion}
To obtain high-order homophily on heterophilous graphs, we propose a path sampling strategy based on feature similarity. This paves a way for a lightweight model named PathMLP, which can encode messages carried by paths via simple transformation and concatenation operations, and effectively learn node representations. Experiments demonstrate that our path sampling strategy is capable of obtain higher-order information, effectively avoiding the over-smoothing problem caused by layer stacking. Meanwhile, all feature extraction, propagation, aggregation, and transformation are implemented through MLP only, making PathMLP highly computationally efficient. Finally, our methods achieves SOTA on 16 out of 20 real-world datasets, underlining its generality across datasets.

In our model, we limit path sampling to the Top-1 node by feature similarity for orders above five to avoid path explosion, and only a subset of paths is randomly selected for training to keep the model lightweight. However, it should be noted that these measures may not always guarantee the effectiveness of selected paths in learning node representations. Therefore, future efforts should focus on achieving a better balance between sampling depth and the number of training paths, as well as exploring more effective path selection or synergizing path selection with model training.

\section*{CRediT authorship contribution statement}
\textbf{Jiajun Zhou:} Conceptualization, Methodology, Writing - Original Draft, Writing - Review \& Editing, Supervision.
\textbf{Chenxuan Xie:} Conceptualization, Methodology, Software, Writing - Original Draft, Data Curation, Investigation.
\textbf{Shengbo Gong:} Conceptualization, Methodology.
\textbf{Jiaxu Qian:} Writing - Original Draft.
\textbf{Shanqing Yu:} Funding acquisition, Supervision.
\textbf{Qi Xuan:} Funding acquisition, Supervision.
\textbf{Xiaoniu Yang:} Funding acquisition, Supervision.

\section*{Declaration of competing interest}
The authors declare that they have no known competing financial interests or personal relationships that could have appeared
to influence the work reported in this paper.

\section*{Data availability}
Data will be made available on request.

\section*{Acknowledgments}
This work was supported in part by the Key R\&D Program of Zhejiang under Grants 2022C01018 and 2024C01025, by the National Natural Science Foundation of China under Grants 62103374 and U21B2001.

\appendix

% \newpage
% \onecolumn
\section{Details of Datasets}\label{app: dataset}
% SNGNN
\subsection{Normal Group}
\begin{itemize}[leftmargin=10pt]
    \item \textit{Cora}, \textit{Citeseer} and \textit{Pubmed}~\cite{cora} are homophily literature citation networks, where nodes represent papers and edges represent the citation relationship between papers. Node features are the bag-of-words which describe whether each word exists in the paper, and node label denotes the research field.
    \item \textit{Cora-full}~\cite{corafull1,corafull2} contains papers from more research fields based on \textit{Cora}.
    \item \textit{Actor}~\cite{actor} is an actor co-occurrence network, where nodes denote actors, edges denote co-occurrence relationships of actors on Wikipedia pages, and node features are constructed from keywords of Wikipedia pages.
    \item \textit{Chameleon-f} and \textit{Squirrel-f}~\cite{evaluate} are the deduplicated versions of \textit{Chameleon} and \textit{Squirrel}, respectively.
    \item \textit{Tolokers}~\cite{evaluate} is a social network from Toloka crowdsourcing platform, where nodes represent workers and two workers are connected if they work on the same task. Node features are constructed from workers' profile information and task performance statistics, and labels indicate whether a worker is banned in a project.
    \item \textit{Roman-empire}~\cite{evaluate} is constructed from the Roman Empire article in Wikipedia, where nodes denote words in the article, edges denote word dependencies, node features are constructed from word embeddings extracted by the FastText method, and labels are the syntactic roles of words.
    \item \textit{Penn94}~\cite{LINKX} is a Facebook social network, where nodes denote students and are labeled with the gender of users, edges represent the relationship of students. Node features are constructed from basic information about students which are major, second major/minor, dorm/house, year and high school.
    \item \textit{Electronics}~\cite{electronics} is an amazon product network, where nodes represent the product of Electronics and edges represent the association of various products. This dataset classifies electronic products into 167 categories, while node features are constructed from product information. 
\end{itemize}

\subsection{Anomaly Group}
\begin{itemize}[leftmargin=10pt]
    \item \textit{Chameleon} and \textit{Squirrel}~\cite{LINKX} are Wikipedia page-page networks, where nodes represent pages of articles and edges represent links of pages. Node features are constructed from keywords in articles.
    \item \textit{Texas}, \textit{Cornell} and \textit{Wisconsin}~\cite{GeomGCN} are heterophilous webpage datasets, where nodes represent webpages and edges represent hyperlinks. Node features are the bag-of-words and node labels are categories of pages.
    \item \textit{NBA}~\cite{NBA} is a social network about NBA basketball players, where nodes denote basketball players and edges denote the relationships between players. Node features are constructed from the statistics information of players, and the players are categorized into U.S. players and oversea players.
    \item \textit{Questions}~\cite{evaluate} dataset collect the question-answering data from website Yandex Q, where nodes represent users and edges represent the link relationships between users. Node features are constructed from embeddings extracted by FastText method, and node labels represent whether users stay active on the website.
    \item \textit{Amazon-ratings}~\cite{evaluate} is an amazon co-purchasing network, where nodes represent products and edges connect products that are frequently co-purchased. This dataset utlizes FastText method to generate embeddings as node features like Questions. The task is to predict the rating values for products in five classes.
    \item \textit{BGP}~\cite{BGP} is a Border Gateway Protocol Network. The BGP network consists of many router nodes, which are connected by BGP sessions. 
\end{itemize}

\subsection{Data Leakage Detection}\label{app: data-leakage}
We design simple methods to detect and verify data leakage. Specifically, for the detection experiments, we first concatenate the adjacency matrix $\boldsymbol{A}$ and label matrix $\boldsymbol{Y}$, then check for the duplication rate of each vector ($\boldsymbol{A}_i \parallel \boldsymbol{Y}_i$); for the verification experiment, we design a two-layer MLP, incorporating features from a mapped adjacency matrix after the first layer, similar to LINKX. If its performance significantly exceeds that of the MLP, it is likely that the model is using topological information for matching rather than node features, i.e., it has learned shortcuts. We detect and verify all datasets, with results displayed in Table~\ref{tab: data-leakage}.
\begin{table}[!htb]
  \centering
  \caption{Results for detecting and verifying data leakage. The top two rows display the results of duplication rate detection, while the bottom three rows display the results of verification experiment (accuracy $\pm$ standard deviation) and performance gain.}
  \label{tab: data-leakage}
  \resizebox{\linewidth}{!}{
  \begin{tblr}{
    cells = {c},
    vline{2} = {-}{dashed},
    hline{1,7} = {-}{0.08em},
    hline{2,4} = {-}{dashed},
  }
                                    & Chameleon    & Squirrel     & Questions    & Amazon-ratings & BGP          \\
  $\boldsymbol{A}$                  & 46.29\%      & 36.42\%      & 42.69\%      & 16.50\%        & 44.86\%      \\
  $\boldsymbol{A} + \boldsymbol{Y}$ & 46.03\%      & 36.28\%      & 42.06\%      & 15.69\%        & 43.03\%      \\
  MLP                               & 48.31 ± 3.69 & 30.55 ± 0.98 & 50.92 ± 1.48 & 44.19 ± 0.61   & 64.41 ± 0.83 \\
  MLP+$\boldsymbol{A}$              & 73.56 ± 3.05 & 65.68 ± 1.48 & 72.18 ± 1.27 & 52.65 ± 0.48   & 64.91 ± 0.60 \\
  gain                              & 52.26\% $\uparrow$  & 114.99\% $\uparrow$ &41.75\% $\uparrow$ &19.14\% $\uparrow$ &0.77\% $\uparrow$
  \end{tblr}}
\end{table}

\section{Details of Hyper-parameters} \label{app: parameter}
\subsection{PathMLP and PathMLP+}
For the transformation dimension $f'$ of $\text{MLP}_1$, we search in \{12, 24, 32\} to avoid the dimensional explosion problem introduced by node feature concatenation.
For the number of paths $N$ introduced to preventing over-squashing, we search in \{2, 4, 6, 8, 10, 12, 15, 18\}.
For the path length parameter $d$, we search in \{3, 4, 5\}.
For the balance coefficient between node features and topological information, we search in \{0, 0.3, 0.5\}, where 0 means no topological information is used.
The search spaces for all parameters are summarized as follows:
\begin{itemize}[leftmargin=10pt]
  \item \textbf{transformation dimension $f'$}: $\{12, 24, 32\}$;
  \item \textbf{number of paths $N$}: $\{2, 4, 6, 8, 10, 12, 15, 18\}$;
  % \item \textbf{dropout}: \{ 0.1, 0.3, 0.5, 0.7, 0.9 \}
  \item \textbf{length parameter of path $d$}: $\{3, 4, 5\}$;
  \item \textbf{balance coefficient $\beta$}: $\{0, 0.3, 0.5\}$;
  \item \textbf{augmentation index $m$}: $\{1, 2\}$;
\end{itemize}

\subsection{Baselines}
% 详细的细节说明
We also use the NNI (Neural Network Intelligence) tuning tool in baselines with the same settings as our methods except for the specific hyper-parameters of baselines. The seaech spaces for specific hyper-parameter are summarized as follow: 
\begin{itemize}[leftmargin=10pt]
    \item \textbf{FAGCN}: $\epsilon \in \{ 0.2, 0.3, 0.4, 0.5 \}$;
    \item \textbf{GPRGNN}: $K \in \{ 10 \}$, dropout $\in \{ 0.5 \}$, $\alpha \in \{ 0.5 \}$;% init $\in \{ \text{PPR} \}$;
    \item \textbf{ACMGCN}: variant $\in \{ \text{False} \}$, is\_need\_struct $\in \{ \text{False} \} $;
    \item \textbf{GCNII*}: $ \alpha \in \{ 0.5 \}, ~\lambda \in \{ 0.5 \}$, variant $\in \{ \text{True} \}$;
    \item \textbf{H2GCN-1}: num\_layers $\in \{ 1 \}$, num\_mlp\_layers $\in \{ 1 \}$;
    \item \textbf{FSGNN}: aggregator $ \in \{ \text{cat}, \text{sum} \}$;
    % \item \textbf{GGCN}: layers $ \in \{ 2, 4, 5, 6, 10 \} $, decay\_rate $\in \{ 0.1, 0.2, 0.3, 0.4, 0.5, 0.6, 0.7, 0.8, 0.9 \}$.
    \item \textbf{GloGNN++}: norm\_layers $ \in \{ 1, 2, 3 \}$, orders $\in \{ 2, 3, 4 \}$, \\
    $\alpha \in \{ 0.0, 1.0 \}$, $\beta \in \{ 0.1, 1.0, 0.05, 800, 1000 \}$, \\
    $\gamma \in \{ 0, 0.1, 0.2, 0.3, 0.4, 0.5, 0.7, 0.8, 0.9 \}$, 
    $\delta \in \{ 0, 0.9, 1.0 \}$;
    \item \textbf{SNGNN+}: $k \in \{ 1, 2, 5, 10, 30, 50 \}$, num\_layers $\in \{ 1, 2 \}$,\\
     $\theta \in \{ 0.99, 0.9, 0.8, 0.5, 0.0, -0.5, -1.0 \}$.
\end{itemize}

\subsection{Common Hyper-parameter for All Methods}
To make a relatively fair comparison, we set the common hyper-parameter search space of all methods to be consistent.
\begin{itemize}[leftmargin=10pt]
  \item \textbf{learning rate}: $\{ 0.005, 0.01, 0.05, 0.1 \}$;
  % \item \textbf{weight decay}: \{5e-5, 1e-5, 5e-4, 1e-4, 5e-3\};
  \item \textbf{dropout}: $\{ 0.1, 0.3, 0.5, 0.7, 0.9 \}$;
  \item \textbf{hidden dimension} $f_h$: $\{ 64 \}$;
\end{itemize}

\printcredits

% \bibliographystyle{cas-model2-names}
\bibliographystyle{elsarticle-num}

\bibliography{acmart}

% % Biography
% % Here goes the biography details.

% % Here goes the biography details.

\end{document}